\pdfoutput=1
\documentclass[11pt]{article}

\usepackage{acl}

\usepackage{times}
\usepackage{latexsym}

\usepackage[T1]{fontenc}
\usepackage[utf8]{inputenc}

\usepackage{microtype}

\usepackage{xspace}
\usepackage[ruled,noline,linesnumbered]{algorithm2e}
\usepackage{algorithm2e}
\usepackage{algorithmic}
\usepackage{microtype}
\usepackage{xcolor}
\usepackage{colortbl}
\usepackage{hhline}
\usepackage{amsmath}
\usepackage{amssymb}
\usepackage{multirow}
\usepackage{booktabs}
\usepackage{graphicx}
\usepackage{enumitem}
\usepackage{tikz}
\usepackage{booktabs}
\usepackage{wrapfig}
\usepackage[normalem]{ulem}
\usepackage{comment}
\usepackage{soul}
\usepackage{tipa}

\title{A$^\textbf{*}$ Decoding: Neural Language Generation with Looking Ahead Heuristic}
\title{A$^\textbf{*}$ Decoding: Neural Language Generation with Look-Ahead Heuristics}
\title{\method: Neural Language Generation with\\Look-Ahead Heuristics}
\title{ A$^\textbf{*}$esque Decoding: \\ Neural Language Generation with Look-Ahead Heuristics  }
\title{NeuroLogic L* Decoding: \\ Constrained Text Generation with Lookahead Heuristics }
\title{\textsc{NeuroLogic L$^*$} Decoding: \\ Constrained Text Generation with Lookahead Heuristics }
\title{\textsc{NeuroLogic L$^\star$} \xspace Decoding: \\ Constrained Text Generation with Lookahead Heuristics }
\title{\textsc{NeuroLogic A$^\bigstar$}{esque} \xspace Decoding: \\ Constrained Text Generation with Lookahead Heuristics }
\author{
\textbf{Ximing Lu}\textsuperscript{$\ddagger$}\textsuperscript{$\dagger$}
\hspace{.1cm}\textbf{$^\heartsuit$Sean Welleck}\textsuperscript{$\dagger\ddagger$} \hspace{.1cm} \textbf{$^\heartsuit$Peter West}\textsuperscript{$\dagger$}\\ \hspace{.1cm}\hspace{.1cm} \textbf{Liwei Jiang}\textsuperscript{$\ddagger\dagger$} \hspace{.1cm} \textbf{Jungo Kasai}\textsuperscript{$\ddagger\dagger$} 
\hspace{.1cm} \textbf{Daniel Khashabi}\textsuperscript{$\ddagger$} 
\hspace{.1cm} \textbf{Ronan Le Bras}\textsuperscript{$\ddagger$}
\hspace{.1cm} \\ \textbf{Lianhui Qin}\textsuperscript{$\dagger$} \hspace{.1cm}
 \textbf{Youngjae Yu}\textsuperscript{$\ddagger$} \hspace{.1cm} \textbf{Rowan Zellers}\textsuperscript{$\dagger$} \hspace{.1cm} \textbf{Noah A.\ Smith}\textsuperscript{$\dagger\ddagger$} \hspace{.1cm} \textbf{Yejin Choi \textsuperscript{$\dagger\ddagger$}}\\
\textsuperscript{$\ddagger$}Allen Institute for Artificial Intelligence\\
  \textsuperscript{$\dagger$}Paul G. Allen School of Computer Science \& Engineering, University of Washington
}

\newcommand{\commongen}{\frenchspacing C{\scriptsize OMMON}G{\scriptsize EN}\xspace}

\newcommand{\commongenfull}{Constrained Commonsense Generation\xspace}
\newcommand{\mtfull}{Constrained Machine Translation\xspace}
\newcommand{\tttfull}{Table-to-text Generation\xspace}
\newcommand{\questiongenfull}{Constrained Question Generation\xspace}
\newcommand{\storygenfull}{Commonsense Story Generation\xspace}

\newcommand{\commongenshort}{Commonsense Generation\xspace}
\newcommand{\mtshort}{Machine Translation\xspace}
\newcommand{\tttshort}{Table-to-text Generation\xspace}
\newcommand{\questiongenshort}{Question Generation\xspace}
\newcommand{\storygenshort}{Commonsense Story Generation\xspace}

\newcommand{\neurologic}{\textsc{NeuroLogic\xspace}}
\newcommand{\neurologicshort}{\textsc{NeuroLogic}\xspace}

\newcommand{\methodnospace}{\textsc{NeuroLogic A$^\bigstar$}esque\xspace}
\newcommand{\method}{\methodnospace{}\xspace}
\newcommand{\astaresque}{\textsc{A$^\bigstar$}esque\xspace}

\newcommand{\methodshortnospace}{\textsc{NeuroLogic$^\bigstar$}}
\newcommand{\methodshort}{\methodshortnospace\xspace}

\newcommand{\astar}{A*\xspace}

\newcommand{\eg}{\textit{e.g.},\xspace}
\newcommand{\ie}{\textit{i.e.},\xspace}

\newcommand{\interalia}[1]{\citep[\emph{inter alia}]{#1}}

\usepackage{amsmath}
\newcommand{\beamwidth}{\ensuremath{k}}
\newcommand{\lk}{\ensuremath{\ell}}

\newcommand{\xb}{\ensuremath{\mathbf{x}}}

\newcommand{\yb}{\ensuremath{\mathbf{y}}}

\DeclareMathOperator*{\argtopk}{arg\,topk} % thin space, limits underneath in displays
\DeclareMathOperator*{\argmax}{arg\,max}

\usepackage{afterpage}
\usepackage{arydshln}

\begin{document}

\maketitle

\renewcommand\thefootnote{}\footnote{$\heartsuit$ Co-second-authors. Other authors are listed alphabetically, as all contributed significantly.}

\renewcommand*{\thefootnote}{\arabic{footnote}}
\setcounter{footnote}{0}

\begin{abstract}

The dominant paradigm for neural text generation is left-to-right decoding from autoregressive language models. 
Constrained or controllable generation under complex lexical constraints, however, requires foresight to plan ahead feasible future paths. 

Drawing inspiration from the \astar search algorithm, we propose \underline{\method},\footnote{pronounced 
\textipa{[ey st\textscripta r  \textepsilon sk]}.} a decoding algorithm 
that incorporates heuristic estimates of future cost.
We develop efficient \textit{lookahead} heuristics that are efficient for large-scale language models, making our method a drop-in replacement for common techniques such as beam search and top-k sampling.
To enable constrained generation, we build on \neurologic~decoding \cite{lu-etal-2021-neurologic}, combining its flexibility in incorporating  logical constraints with \astaresque estimates of future constraint satisfaction.

Our approach outperforms competitive baselines on five generation tasks, and achieves new state-of-the-art performance on table-to-text generation, constrained machine translation, and keyword-constrained generation. 
The improvements are particularly notable on tasks that require complex constraint satisfaction or in few-shot or zero-shot settings. 
%which benefits from the planning ahead into the future. 
\method illustrates the power of decoding for improving and enabling new capabilities of large-scale language models.
% The effectiveness of our approach motivates further research on inference algorithms for constrained neural text generation.  
% The effectiveness of \method motivates the needs for future research to enhance inference algorithms for constrained neural text generation.  
%sheds light on the pressing need for models and algorithms that make globally optimal decisions, considering past and future text together. 

% \footnote{pronounced \textit{ay-star-esk}}
% Inspired by guided search algorithms, we combine beam search with \textit{look-ahead heuristics} 
% to form an expectation of future generation states.
% This guided generation via lookahead heuristics guides the output generations 
% towards those that satisfy logical constraint while maintaining a  fluent and coherent language. 
% practical efficient heuristics 
% 2nd point: combining it with constraint generation 

\begin{comment}
Inspired by the \astar \danielchange{graph traversal} 
% \st{search}  
algorithm, we propose \method\footnote{pronounced \textit{ay-star-esk}} (shortened as \methodshortnospace), to allow current decoding algorithms to consider the \emph{future} along with the past. 
\methodshort uses \textit{look-ahead heuristics} 
\danielchange{to form an expectation of future generation states.
This guided generation via lookahead heuristics guides the output generations 
towards those that satisfy logical constraint while maintaining a  fluent and coherent language. }
\end{comment}

\end{abstract}

\section{Introduction}

\begin{figure}
    \centering
    \includegraphics[width=1\linewidth]{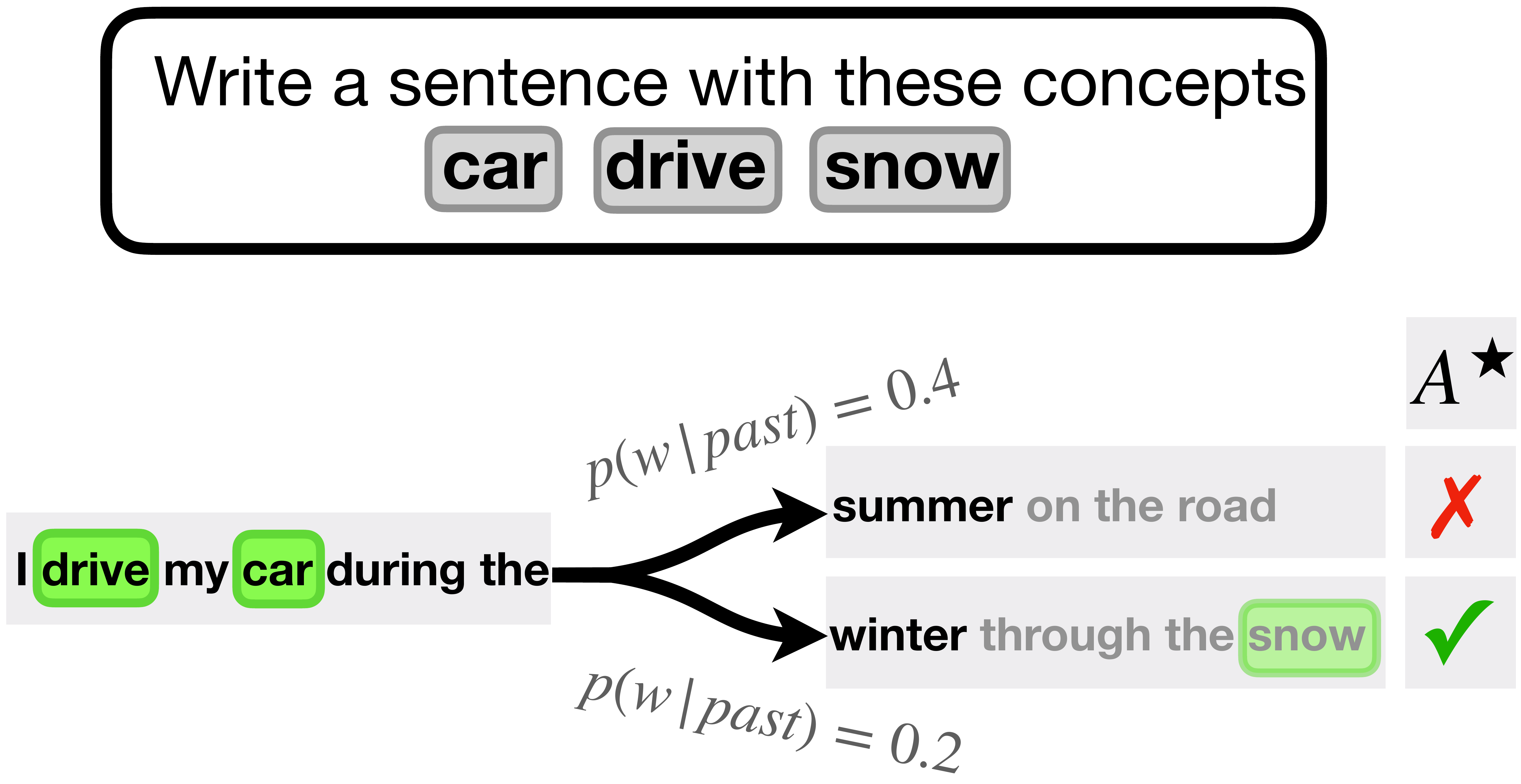}
    \caption{
        \methodshort leverages \textit{lookahead heuristics} to guide generations towards those that satisfy the given task-specific constraints. 
        %  global constraints. \ximing{those that are more consistent with future constraints} 
        In this example from the \commongen task, although \textbf{summer} is a \emph{more likely} next word given the already-generated \emph{past}, \methodshort looks ahead to see that selecting \textbf{winter} results in a generation that incorporates unsatisfied constraint \textbf{snow} with a higher probability later on. Thus, \textbf{winter} is preferred despite being lower probability than \textbf{summer}.
    }
    \label{fig:teaser}
\end{figure}

The dominant paradigm for neural text generation is based on left-to-right decoding from autoregressive language models such as GPT-2/3 \cite{radford2019language,Brown2020LanguageMA}. 
% Under this paradigm, a common characteristic underlying various decoding techniques such as beam search or top-k/p sampling   \cite{holtzman2020curious} is that the decision to generate a token at time $t$ primarily reflects what happened in the past, without explicitly looking ahead into the future. 
Under this paradigm, common decoding techniques such as beam search or top-k/p sampling \cite{holtzman2020curious} determine which token to generate next based on what happened in the past, without explicitly looking ahead into the future. 
While this lack of foresight often suffices for open-ended text generation -- where any coherent text can be acceptable -- for constrained text generation, planning ahead is crucial for incorporating all desired content in the generated output \cite{hu2017toward,dathathri2019plug}.

Classical search algorithms such as A* search  \cite{hart1968,Pearl1984HeuristicsI,korf1985depth} address the challenge of planning ahead by using \textit{heuristic} estimation of future cost when making decisions. 
Drawing inspiration from A* search, we develop \method (shortened to \methodshort), which combines A*-like heuristic estimates of future cost (e.g. perplexity, constraint satisfaction) with common decoding algorithms for neural text generation (e.g. beam search, top-$k$ sampling), while preserving the efficiency demanded by large-scale neural language models. % while determining which token to generate next.
% where the cost is the perplexity of the generated text optionally combined with a penalty to guide constraint satisfaction (e.g. incorporating desired facts into table-to-text generation). 
% For example, an estimated future cost that guides the model to generate outputs consistent with the given structured information for table-to-text generation. 

As selecting the next token to generate based on the \textit{optimal} future cost is NP-complete \cite{chen-etal-2018-recurrent}, 
we develop \textit{lookahead} heuristics, which approximate cost at each decoding step based on continuations of the sequence-so-far. %partially decoded sequences.
% We draw on classical search algorithms such as A* search \cite{Pearl1984HeuristicsI,korf1985depth,hart1968}, which improve search by planning ahead using \textit{heuristic} estimates of future cost when making each decision. %when making the current decision
% We develop \method (shortened to \methodshort), which flexibly incorporates estimate of future cost (e.g. perplexity, constraint satisfaction) for decoding with modern neural language models.
%
% Unlike traversal graphs considered in the classical AI literature however, estimating future cost through autoregressive language models poses a unique computational challenge \cite{chen-etal-2018-recurrent} \yejin{can we cite some theoretical papers here, maybe Kevin Knight has one among others?} Therefore, we propose several lookahead-based  heuristics to approximate the future loss based on continuations of the partially decoded sequence. 
%\peter{FROM NOAH: more explanation as to how we relate to astar} Search paths are biased towards ``future gains'', such as overall high probability, and satisfaction of constraints and controls. 
Figure~\ref{fig:teaser} shows an example, where \method guides generation towards a decision that would have been ignored based on the past alone, but is selected after looking ahead and incorporating the probability that constraints are satisfied in the future.
% less likely given the past alone, but with a higher probability of satisfying necessary constraints in the future. 

Our approach builds on \textsc{NeuroLogic} Decoding of \citet{lu-etal-2021-neurologic}, a variation of beam-search for controlling generation through rich logic-based lexical constraints expressed in Conjunctive Normal Form (CNF). 
% decoding a text that satisfies a complex set of constraints expressed in Conjunctive Normal Form (CNF) while also optimizing for the probability of the generated text to ensure fluency. 
% This framework allows for . 
Our work generalizes \citet{lu-etal-2021-neurologic} by (1) incorporating novel lookahead heuristics to estimate future contraint satisfaction, 
%(2) supporting top-k sampling when diversity is desired in the output, 
and (2) developing additional \emph{unconstrained} variants that can work with an empty set of constraints. These new algorithm variants support broad applications of \methodshort, including unconstrained generation, as demonstrated in our experiments. 
%including machine translation or open-ended story continuation, which were not possible to tackle with \neurologic. 

Extensive experiments across five generation tasks demonstrate that our approach outperforms competitive baselines. We test \methodshort in conjunction with both supervised and unsupervised models and find that the performance gain is pronounced especially in zero-shot or few-shot settings. 
% \danielchange{
In particular, 
on the \commongen benchmark, using our proposed decoding algorithm with an off-the-shelf language model outperforms a host of \emph{supervised} baselines with conventional decoding algorithms. % -- i.e. without any labeled data and only through effective incorporation of task constraints. 
% }
% \yejin{should mention when unsupervised neurologic wins over supervised conventional search if at all?}
% \daniel{added a sentence. Someone check?}
This demonstrates that a strong inference-time algorithm such as \methodshort can alleviate the need for costly datasets that are manually annotated for explicit supervision. 
Moreover, we find that \methodshort achieves state-of-the-art performance in various settings, including WMT17 English-German machine translation with lexical constraints \cite{dinu-etal-2019-training} and few-shot E2ENLG table-to-text generation \cite{chen-etal-2020-kgpt}. 

In summary, we develop \methodnospace, a new decoding algorithm for effective and efficient text generation. 
To our knowledge this is the first A*-like algorithm for guided text generation via  lookahead heuristics. 
Our algorithm is versatile, as it can be applied to a variety of tasks via inference-time constraints, reducing the need for costly labeled data. 
Extensive experiments show its effectiveness on several important generation benchmarks.

\section{\method Decoding %\daniel{\st{with Lookahead}}
}
\label{sec:method}
We describe \method Decoding (shortened as \methodshort), our decoding algorithm motivated by A$^*$ search \citep{hart1968},
a best-first search algorithm that finds high-scoring paths using a heuristic estimate of future return.
We first introduce the decoding problem, and then describe our heuristics with a novel lookahead procedure for adapting \methodshort search to unconstrained and constrained generation with large-scale 
% sequence 
autoregressive models.

\subsection{Decoding With \astaresque Lookahead}
\label{ssec:decoding}
\paragraph{Decoding.}
Sequence-to-sequence generation is the task of generating an output sequence $\yb$ given an input sequence $\xb$. 
We consider standard left-to-right, autoregressive models, 
$p_\theta(\yb|\xb) = \prod_{t=1}^{|\yb|} p_\theta (y_t| \yb_{<t},\xb)$, and
omit $\xb$ to reduce clutter.
Decoding consists of solving,
\begin{align}
\label{eqn:decode}
\yb_* = \argmax_{\yb\in \mathcal{Y}} F(\yb).
\end{align}
Where $\mathbf{\mathcal{Y}}$ is the set of all sequences.
In our setting, the objective $F(\yb)$ takes the form $s(\yb)+H(\yb)$, where $s(\yb)$ is $\log p_\theta(\yb)$, and  $H(\yb)$ is either zero or is a score for satisfying constraints on $\yb$.

Our method takes the perspective of decoding as discrete search, in which states are partial prefixes, $\yb_{<t}$, actions are tokens in vocabulary $\mathcal{V}$ (i.e. $y_t\in \mathcal{V}$) and transitions add a token to a prefix, $\yb_{<t}\circ y_t$.
Each step of decoding consists of 1) expanding a set of candidate next-states, 2) scoring each candidate, and 3) selecting the $\beamwidth$ best candidates: 
% \peter{FROM NOAH: confusion about $Y_{t-1}$ and $Y_{t}'$. See his notes}\sean{$Y_t'$ is expanded. $Y_t$ is size $k$.} \yejin{this looked fine for me, though i agree it takes a bit of time to parse out the def. we can revisit this for the real submission if we can simplify?}\sean{sure -- agreed!}
\begin{align}
\label{eqn:update}
    \nonumber Y_{t}' &= \{\yb_{<t}\circ y_t\ |\ \yb_{<t}\in Y_{t-1}, y_t\in \mathcal{V}\},\\
    Y_{t}  &=\argtopk_{(\yb_{<t},y_t)\in Y_{t}'}\ \left\{ f(\yb_{< t}, y_t)\right\},
\end{align}
where $Y_0 = \{\langle bos \rangle\}$ and $f(\cdot)$ is a scoring function that approximates the objective $F$.
Common decoding algorithms such as beam search score candidates
without considering future tokens, e.g., $f(\yb_{<t}, y_t)=\log p_\theta(\yb_{\leq t})$.

\paragraph{Lookahead heuristics.}
Our method incorporates an estimate of the future into candidate selection.
Ideally, we want to select candidates that are on optimal trajectories, replacing \autoref{eqn:update} with:
\begin{align}
\label{eqn:update2}
    Y_t=\argtopk_{(\yb_{<t},y_t)\in Y_{t}'}\ \left\{ \max_{\yb_{>t}} F(\yb_{<t},y_t,\yb_{>t}) \right\}.
\end{align}
% \daniel{the $k$ in "topk" might be confused with the $k$ length of lookahead.}  
% where, 
% \begin{align}
% \label{eqn:qfunc}
% Q(\yb_{<t},y_t)=\max_{\yb_{>t}} V(\yb_{<t},y_t,\yb_{>t}).
% \end{align}
However, computing \autoref{eqn:update2} presents two difficulties: 1) the objective $F(\yb)$ may be unknown or difficult to compute, and 2) the space of future trajectories $\yb_{>t}$ is prohibitively large.

Motivated by A$^*$ search~\citep{hart1968}, a best-first search algorithm that finds high-scoring paths by selecting actions that maximize,
$$f(a)=s(a)+h(a),$$ where $s(a)$ is the score-so-far  and $h(a)$ is a heuristic estimate of the future score.
We approximate the objective using a lightweight \textit{heuristic} $h(\cdot)$,
\begin{align}
\label{eqn:update3}
    Y_t=\argtopk_{\yb_{\leq t}\in Y_{t}'}\ \left\{ s(\yb_{\leq t}) + \max_{\yb_{>t}} h(\yb_{<t},y_t,\yb_{>t}) \right\},
\end{align}
where $s(\yb_{\leq t})=\log p_\theta(\yb_{\leq t}).$
To make the search tractable, we search over a set of \textit{lookahead} continuations, approximating \autoref{eqn:update2} as, 
\begin{align}
\label{eqn:update4}
    Y_t=\argtopk_{\yb_{\leq t}\in Y_{t}'}\ \left\{ s(\yb_{\leq t}) + \max_{\mathcal{L}_\lk(\yb_{\leq t})} h(\yb_{\leq t+\lk}) \right\},
\end{align}
where each element $\yb_{t+1:t+\lk}$ of  $\mathcal{L}_{\lk}(\yb_{\leq t})$  is a length-$\lk$ continuation of $\yb_{\leq t}$.
Beam search corresponds to setting $\lk$ and $h$ to $0$.

\paragraph{\astaresque decoding.}
Beam search, \astar search, and our method fall under a general class of algorithms that differ based on (1) which candidates are expanded, (2) which candidates are pruned, (3) how candidates are scored \citep{meister2020best}.
We inherit the practical advantages of beam search-style expansion and pruning, while drawing on \astar-like heuristics to incorporate estimates of the future, and 
% Our approach to expanding and pruning is adopted from beam-search -- expanding $k$ candidates of the current length and selecting the top-$k$ expansions -- while \astar traditionally uses best-first expansion without pruning.
% However, our approach scores candidates with a heuristic, similar to \astar.
% Our heuristics use an explicit lookahead, meaning that our method is summarized as beam search with a lookahead heuristic. 
refer to our method as \textbf{\astaresque decoding}. 
%, as our method draws on A$^*$-like heuristics while favoring the practical advantages of beam search-style expansion and pruning.

\paragraph{Generating lookaheads.}
We compare several methods for generating the lookaheads $\mathcal{L}_\lk(\yb_{\leq t})$.

The \textit{greedy} lookahead produces a single sequence, $\mathcal{L}_\lk=\{\yb_{t+1:t+\lk}\}$, starting from $\yb_{\leq t}$ and selecting each token according to
$y_{t'} =\argmax_{y\in \mathcal{V}} p_\theta(y|\mathbf{y}_{< t'})$.

We also consider a relaxation which interpolates between providing the greedy token and a uniform mixture of tokens as input at each step. 
Specifically, we adjust the model's probabilities with a temperature, $\tilde p_\theta(y_t|\yb_{<t})=\text{softmax}(s_t/\tau)$, where $s_t\in \mathbb{R}^{|\mathcal{V}|}$ is a vector of logits,
and feed the expected token embedding as input at step $t$,
\begin{align}
e_t=\mathbb{E}_{y_t\sim \tilde p(y_t|\yb_{<t})}[E(y_t)],
\end{align}
where $E\in \mathbb{R}^{\mathcal{|V|}\times d}$ is the model's token embedding matrix.
This \textit{soft lookahead} moves from providing the greedy token as input ($\tau\rightarrow 0$) to a uniform mixture of tokens ($\tau \rightarrow \infty$) based on the value of temperature $\tau$.
When using the soft lookahead, we use $\tilde p$ in place of $p$ when scoring tokens.
The soft (and greedy) lookahead is efficient, but only explores a single trajectory.

The \textit{beam} lookahead trades off efficiency for exploration, returning a set $\mathcal{L}_\lk$ containing the top-$\beamwidth$ candidates obtained by running beam search for $\lk$ steps starting from $\yb_{<t}$.

Finally, the \textit{sampling} lookahead explores beyond the highly-probable beam search continuations, generating  each $\yb_{t+1:t+\lk}\in \mathcal{L}_\lk$ using,
\begin{align*}
    y_{t'} \sim p_\theta(y|\mathbf{y}_{< t'}),
\end{align*}
for $t'$ from $t$+1 to $t$+k.

%
% the correspondence with A$^*$ search implies that if the heuristic $h(\yb_{\leq t+\lk})$ is admissible, meaning that the heuristic never overestimates future return, and the beam width is infinite, lookahead decoding will return an optimal sequence.
% However, these guarantees do not apply for practical beam sizes. 
%
% More importantly, practical decoding algorithms for large-scale language models are \textit{approximate}.
% Instead of seeking optimality, our focus is thus on 
% and our goal is to develop an approximate search that improves over beam search. Rather than Thus, we 
% However, using an admissible heuristic is a sufficient, but not necessary, condition.
% evaluate our lookahead heuristics empirically, and save further theoretical analysis as future work.
% \fi
%
Next, we move to our proposed lookahead heuristics, starting with the unconstrained setting.

\subsection{Unconstrained Generation with \methodshort}

First we consider a standard decoding setting, 
$$\argmax_{\yb\in \mathcal{Y}} \log p_\theta(\yb|\xb).$$
We score candidates based on a combination of the \textit{history} and \textit{estimated future}, by using the likelihood of the lookahead as a heuristic. 
% \liwei{by applying the lookahead heuristic}\sean{we still need to define what the lookahead heuristic is in this unconstrained case}
That is, at the $t$th step of decoding, we use \autoref{eqn:update4}:
\begin{align}
\label{eqn:heur-unc}
    h(\yb_{\leq t+\lk}) &= \lambda \log p_\theta(\yb_{t+1:t+\lk}|\yb_{\leq t}, \xb),
\end{align}
where $\lambda$ controls how much we rely on the estimated future versus the history, similar to weighted \astar \cite{pohl1969first}.

\subsection{\methodshort for Constrained Generation}
% Lexically-constrained generation is the task of generating text subject to constraints on which tokens should or shouldn't appear.
Our lookahead heuristics lend themselves to decoding with lexical constraints in a way that standard beam search does not.
For constrained generation, we build on and generalize \neurologicshort decoding algorithm of  \citet{lu-etal-2021-neurologic}-- a beam-based search algorithm that supports a wide class of logical constraints for lexically constrained generation-- with estimates of future contraint satisfaction.

\paragraph{Background of \neurologicshort.} 
\neurologicshort \citet{lu-etal-2021-neurologic} accepts lexical constraints in Conjunctive Normal Form (CNF):
\[\underbrace{\big(D_1 \lor D_2 \cdots \lor D_i \big)}_{C_1}\land \cdots \land \underbrace{\big(D_{i'} \lor D_{i'+1} \cdots \lor D_N \big)}_{C_M} \]
where each $D_i$ represents a single positive or negative constraint,  $D(\textbf{a}, \textbf{y})$ or $\neg D(\textbf{a}, \textbf{y})$, enforcing the phrase $\textbf{a}$ to be included in or omitted from \textbf{y}. \citet{lu-etal-2021-neurologic} refer to each constraint $D_i$ as a \emph{literal}, and each disjunction $C_j$ of literals as a \emph{clause}. %\methodshort seeks optimal sequences in which all clauses are satisfied.

\neurologicshort is a beam-based approximate search for an objective which seeks fluent sequences in which all clauses are satisfied:
\vspace*{-3mm}
\begin{align*}
    \arg\max_{\textbf{y} \in \mathcal{Y}}p_\theta(\textbf{y}|\textbf{x}) - \lambda' \sum_{j=1}^{M} (1-C_j),
 \end{align*}
where $\lambda'\gg 0$ penalizes unsatisfied clauses.
At each step of the search, \neurologicshort scores each of the $k\times |\mathcal{V}|$ candidates $(\yb_{<t},y_t)$ based on whether they (partially) satisfy new constraints,
\begin{align}
\label{eqn:neurologic-score}
    f(\yb_{\leq t}) = &\> \log p_\theta(\textbf{y}_{\leq t}|\xb) + \lambda_1 \max_{\substack{D(\textbf{a},\yb_{\leq t})}} \frac{|\hat{\textbf{a}}| }{|\textbf{a}|},
\end{align}
% \peter{FROM NOAH: $\substack{D(\textbf{a},\yb_{\leq t})}$ needs scope (i.e. what is this over?) ALSO fix this in equation (8)}
% \jungo{I think text below explains that. I think math notations are tricky, so this is fine?}
% \sean{Yes -- we could clean it up for the Jan15 version, we gave it a shot earlier but it was tricky}
where the maximization is over a set of unsatisfied multi-token constraints $\textbf{a}$ tracked by \neurologicshort,  and $\hat{\textbf{a}}$ is the prefix of $\textbf{a}$ in the ongoing generation.
For example, for $\yb_{\leq t}=$``\textit{The boy climbs an apple}'' and constraint $\textbf{a}$=``\textit{apple tree}'', $\hat{\textbf{a}}$ is ``\textit{apple}''.
Intuitively, this function rewards candidates that are in the process of satisfying a constraint.

In lieu of taking the top-$k$ scoring candidates (\autoref{eqn:update4}), \neurologicshort prunes candidates that contain clauses that violate constraints, groups the candidates to promote diversity, and selects high-scoring candidates from each group. 
We use the same pruning and grouping approach, and refer the reader to \citet{lu-etal-2021-neurologic} for further details.
% At each time step, \methodshort chooses which candidates to expand next by 3 steps: \emph{pruning}, \emph{grouping}, and \emph{selecting} 

% We generalize \citep{lu-etal-2021-neurologic} by 

% \citet{lu-etal-2021-neurologic} propose NeuroLogic, which adjusts a candidate $(\yb_{<t},y_t)$'s scores based on whether the new token $y_t$ satisfies a logical constraint:
% \vspace*{-3mm}
% \begin{align*}
% \hat V = &\> \log p_\theta(y_t| \textbf{y}_{<t},\xb) + \lambda_1 \cdot \max_{\substack{D(\textbf{a}_\textbf{i}, \textbf{y}) \\ \in \> \textnormal{state S1}}} \frac{|\hat{\textbf{a}}_\textbf{i}| }{|\textbf{a}_\textbf{i}|}.
%  \end{align*}
% Here, \textcolor{red}{[TODO: concise explanation of the above. probably need to adjust the notation too (e.g. maybe we don't need state S1 here)]}. \ximing{Intuitively, this score function ranks candidates by log likelihood and gives a partial reward to the candidates in the process of satisfying a multi-tokens positive constraint.}
% \citet{lu-etal-2021-neurologic} show that using beam search with this scoring function empirically approximates,
% \vspace*{-3mm}
% \begin{align*}
%     \arg\max_{\yb \in \mathcal{Y}}P_\theta(\yb|\xb) - \lambda' \sum_{i=1}^{L} (1-C_i).
%  \end{align*}
% See the Appendix for a full review of NeuroLogic.

\paragraph{\methodshort decoding.}
Our method improves upon the \neurologicshort scoring function with an estimate of future constraint satisfaction.
Our key addition is a lookahead heuristic that adjusts a candidate $(\yb_{<t},y_t)$'s score proportional to the probability of satisfying additional constraints in the lookahead $\yb_{t+1:t+\lk}$:
\begin{multline}
    h_{\text{future}}(\yb_{\leq t+\lk})= \\ \lambda_2 \max_{D(\mathbf{a},\yb_{\leq t})} \log p_\theta(D(\mathbf{a},\yb_{t+1:t+\lk})|\xb,\yb_{\leq t}) \label{astar_heuristic},
\end{multline}
where we define the probability that constraint $\mathbf{a}$ is satisfied using the most probable subsequence,
\begin{multline}
    p_\theta(D(\mathbf{a},\yb_{t+1:t+\lk})|\xb,\yb_{\leq t}) = \\
        \max_{t'\in [t,t+\ell]} p_\theta(\yb_{t':t'+|\mathbf{a}|}=\mathbf{a}|\xb,\yb_{<t'}) \label{astar_constraint_score},
        % \max_{i\in [1,\lk]}[z_1,\ldots,z_\ell],
\end{multline}
% where $z_i=\log p_\thet$
$\lambda_2$ is a scaling hyperparameter for the heuristic.
% \ronan{Did you want to introduce $\lambda_2$ in (\ref{astar_heuristic})?} \ronan{In (\ref{astar_constraint_score}), what is $i$?}% \liwei{for what? maybe elaborate (e.g., a scaling hyperparameter for ...)}.
% \sean{fixed}

% \jungo{Should be $\yb_{t+i:t+i+|\mathbf{a}|} = \mathbf{a}$ (right hand side)?}

Intuitively, this lookahead heuristic brings two benefits.
When $y_t$ is a token that would satisfy a multi-token constraint, the lookahead incorporates the score of the \textit{full} constraint. 
% \ximing{in this case, we also compute the later score in the following sentence, and take max over the two cases} \sean{can you clarify?}
When $y_t$ is a token that is not part of a constraint, the lookahead allows for incorporating the score of a future constraint that would be satisfied if $y_t$ was selected.

We add our lookahead heuristic to the \neurologicshort scoring function (\autoref{eqn:neurologic-score}), and call the resulting decoding procedure \method (or, \methodshort in short).

% \input{section/2_method}
% \afterpage{\blankpage} 
% \newpage
\begin{table}[ht]
    \centering
    \footnotesize
    \setlength{\tabcolsep}{12pt}
    \centering
            \scalebox{0.9}{\begin{tabular}{l @{\hspace{0.5\tabcolsep}} c @{\hspace{0.5\tabcolsep}} c}
        \toprule 
        \hspace{14mm} \textbf{Task} & \textbf{Supervision} & \textbf{Constraints} \\
         \midrule 
        %  \commongen & zero + full & w/ \\ 
        %  Question Gen. & zero & w/ \\ 
        %  Table2Text & few & w/ \\ 
        %  Machine Translation & full & w/ \\ 
        %  RocStories & full & w/o \\ 
        
        \commongenshort & zero+full & w/ \\ 
        \mtshort & full & w/ \\ 
        \tttshort & few & w/ \\ 
        \questiongenshort & zero & w/ \\ 
        \storygenshort & full & w/o \\ 
        \bottomrule
    \end{tabular}}
    
    \caption{
        Tasks and setups considered in this work. 
    }
    \label{tab:experiments:setups}
\end{table}

%\newcolumntype{x}[1]{%
%>{\centering\hspace{0pt}}p{#1}}%
%\newcolumntype{?}{!{\vrule width 1pt}}
\begin{table*}[t]
\setlength{\tabcolsep}{4.0pt}
    \centering
    \renewcommand{\arraystretch}{1.15}
    \scalebox{.7}{
    \begin{tabular}{p{5.0cm}| c c c c c  c |c c c c}
    \specialrule{1.5pt}{-1.5pt}{0pt}
\multirow{2}{*} {\hspace{13mm}\textbf{Decode Method}} & \multicolumn{6}{c}{\textbf{Automatic Evaluation}}  &  \multicolumn{4}{|c}{\textbf{Human Evaluation}} \\ 
 & \textbf{ROUGE-L} &  \textbf{BLEU-4}  & \textbf{METEOR} &\textbf{CIDEr} & \textbf{SPICE} &\textbf{Coverage} & \textbf{Quality} & \textbf{Plausibility} & \textbf{Concepts} & \textbf{Overall} \\ 
%\specialrule{\heavyrulewidth}{-\heavyrulewidth}{0pt}
\specialrule{1.0pt}{-1.0pt}{0pt}
\rowcolor[gray]{0.90} \multicolumn{11}{l}{\textit{Supervised}}
\\
CBS \cite{anderson-etal-2017-guided} & 38.8  & 20.6  & 28.5 & 12.9 & 27.1 & 97.6 & 2.27 & 2.35 & 2.51 & 2.23  \\
GBS \cite{hokamp-liu-2017-lexically} & 38.2 & 18.4 & 26.7 & 11.7 & 26.1 & 97.4 & 2.06 & 2.17 & 2.29 & 2.01 \\
DBA \cite{post-vilar-2018-fast} & 38.3 & 18.7 & 27.7 & 12.4 & 26.3 & 97.5 &2.23 & 2.30 & 2.43 & 2.15  \\
\textsc{NeuroLogic} \cite{lu-etal-2021-neurologic} & 42.8 & 26.7 & 30.2 & 14.7 & 30.3 & \underline{97.7} & 2.54 & 2.56 & 2.67 & 2.50 \\
\hdashline
\methodshort (greedy) & \underline{43.6} & \underline{28.2} & \textbf{30.8} & 15.2 & \underline{30.8} & \textbf{97.8} & 2.66 & \underline{2.67} & 2.73 & 2.59\\
\methodshort (sample) & 43.4 & 27.9 & \textbf{30.8} & \underline{15.3} & \textbf{31.0} & \underline{97.7}& 2.64 & 2.64 & 2.74 & 2.58 \\
\methodshort (beam) & 43.2 & \underline{28.2} & \underline{30.7} & 15.2 & \textbf{31.0} & 97.6 & \underline{2.68} & \underline{2.67} & \underline{2.76} & \underline{2.60} \\

\specialrule{1.0pt}{-1.0pt}{0pt}

\rowcolor[gray]{0.90} \multicolumn{11}{l}{\textit{Unsupervised}}\\
TSMH \cite{zhang-etal-2020-language-generation} & 24.7 & \phantom{0}2.2 & 14.5 & \phantom{0}3.6 & 15.4 & 71.5 & 1.85 & 1.92 & 1.95 & 1.63 \\
\textsc{NeuroLogic} \cite{lu-etal-2021-neurologic} & 41.9 & 24.7 & 29.5 & 14.4 & 27.5 & 96.7 & 2.64 & 2.52 & 2.68 & 2.50\\
\hdashline
\methodshort (greedy) & \textbf{44.3} & \textbf{28.6} & \underline{30.7} & \textbf{15.6} & 29.6 & 97.1 & \textbf{2.78} & \textbf{2.70} & \textbf{2.77} & \textbf{2.70} \\
    \specialrule{1.5pt}{-1.5pt}{0pt}
    \end{tabular}}
    \caption{Performance of various decoding methods with \textit{supervised} or \textit{off-the-shelf} GPT-2 on the \commongen test set, measured with automatic and human evaluations. We only tried \methodshort (greedy) in the unsupervised setting because of the computational cost. The best numbers are \textbf{bolded} and the second best ones are \underline{underlined}.}

    \label{tab:comGen_result_decode}
\end{table*}

\section{Experiments: Constrained Generation}
%\liwei{This is not a priority but maybe we could include detailed experimental set up ((hyper) parameters etc.) in the Appendix}
%\jungo{+1!}

We present experimental results on various constrained generation benchmarks: \commongen (\S\ref{sec:commongen}), constrained machine translation (\S\ref{sec:mt}), table-to-text generation (\S\ref{sec:table-to-text}), and interrogative sentence generation (\S\ref{sec:interrogative}).
\methodshort consistently outperforms \neurologic{} and all previous approaches. The improvement is especially substantial in zero-shot and few-shot cases where the search problem is much harder.
%\liwei{Can we add a short overview paragraph to introduce the tasks, results and main findings?}

\paragraph{Experimental setups.}
We explore a variety of experimental setups (Table~\ref{tab:experiments:setups}).
In terms of \emph{supervision}, we consider different configurations of zero-shot, few-shot and full-shot. 
The former two supervision regimes are particularly important as many realistic generation application do not come with many manually-annotated labeled data.  
Additionally, we study both \emph{constrained} and \emph{unconstrained} tasks, even though we focus on the former.

\paragraph{Evaluation metrics.}
We use the following automatic metrics that are commonly used for evaluating text generation: 
BLEU~\cite{papineni2002bleu}, ROUGE~\cite{lin2004rouge}, METEOR~\citep{banerjee2005meteor},
CIDEr~\citep{vedantam2015cider}, SPICE~\citep{anderson2016spice} and 
NIST~\cite{lin-hovy-2003-automatic}.
Any other domain specific metrics are detailed in each task description. 

\subsection{\commongenfull}
\label{sec:commongen}
% \liwei{Can we provide a table of few concrete examples comparing generations with and without A*?}\ximing{TODO}

\begin{table}[t]
\scriptsize
\centering
    \begin{tabular}{@{}l @{\hspace{0.48\tabcolsep}} l @{\hspace{0.48\tabcolsep}} l @{}}
        \toprule 

        \textbf{Words} & \hspace{4mm}\textbf{Method} &  \hspace{16mm} \textbf{Generation} \\
        \midrule
        
        % show & GBS & Sew machine is used to show how to use the machine. \\
        % machine & DBA & Use a sewing machine to show how to sew a garment. \\
        % sew & \textsc{NL} & A woman uses a sewing machine to show how it works. \\
        % use & \nlastar & \textbf{A man show how to use a sewing machine.} \\
        % \midrule
        
        cut & GBS & Cut a piece of wood to use as a fence. \\
        piece & DBA & Cut a piece of wood to use as a fence. \\
        use & \neurologic & Piece of wood used for cutting. \\
        wood & \methodshort & \textbf{A man cuts a piece of wood using a circular saw.} \\
        \midrule
        
        ball & GBS & A dog is run over by a ball and mouth agape. \\
        dog & DBA & A dog is run over by a ball and bites his mouth. \\
        mouth & \neurologic & A dog is running and chewing on a ball in its mouth. \\
        run & \methodshort & \textbf{A dog running with a ball in its mouth.} \\
        \midrule
        
        dog & GBS & Soap and water scrubbed dog with a towel. \\
        scrub & DBA & Soap and water on a dog and scrubbed skin. \\
        soap & \neurologic & A dog is scrubbing his paws with soap and water. \\
        water & \methodshort & \textbf{A man is scrubbing a dog with soap and water.} \\
        % \midrule
        
        %  & GBS &  \\
        %  & DBA &  \\
        %  & \neurologic &  \\
        %  & \methodshort &  \\

        \bottomrule
    \end{tabular}
    \caption{Example generations for the \commongen task across supervised \methodshortnospace and baselines, including GBS \cite{hokamp-liu-2017-lexically}, DBA \cite{post-vilar-2018-fast}, and \textsc{NeuroLogic} \cite{lu-etal-2021-neurologic}}
\label{tab:commongen_ex}
\end{table}

% \cite{hokamp-liu-2017-lexically}
% \cite{post-vilar-2018-fast}
% (\textsc{NL}\bigstar)

\commongen \cite{lin2019commongen} 
is a constrained commonsense generation task with lexical constraints. 
% that assesses the ability of generative commonsense reasoning.
Given a set of concepts (\eg \{throw, run, javelin, track\}), the task is to generate a coherent sentence describing a plausible scenario using all of the given concepts (\eg ``a man runs on a track and throws a javelin.''). 

\paragraph{Approach and Baselines.}
Following \citet{lu-etal-2021-neurologic}, we enforce that each given concept $c_i$ must appear in output $\textbf{y}$ under some morphological inflection.
We experiment with both supervised and zero-shot settings. 
In the supervised setting, we formulate it as conditional sentence generation task and finetune GPT-2  \cite{radford2019language} as a sequence-to-sequence model. 
In the zero-shot setting, we use GPT-2 off-the-shelf (no fine-tuning), and rely on constrained decoding to guide the generations. We compare with previous constrained decoding algorithms, including CBS \cite{anderson-etal-2017-guided}, GBS \cite{hokamp-liu-2017-lexically}, DBA \cite{post-vilar-2018-fast}, \neurologicshort \cite{lu-etal-2021-neurologic} and TSMH \cite{zhang-etal-2020-language-generation}

%\ximing{add a few lines for baselines}
\paragraph{Metrics} Following \citet{lin2019commongen}, we report automatic generation metrics
% , including BLEU \cite{papineni2002bleu}, ROUGE \cite{lin2004rouge}, METEOR \citep{banerjee2005meteor}, CIDEr \citep{vedantam2015cider}, SPICE \citep{anderson2016spice}, and also 
as well as \textit{coverage}, defined as  the average percentage of the provided concepts that are present  in lemmatized outputs. 
Additionally, we conduct human evaluation on $100$ test examples with workers from Amazon Mechanical Turk (AMT).
We include our evaluation template in Figure \ref{fig:commongen_human_eval} of Appendix \ref{sec:appendix:human_eval}. 
Workers are given a pair of concepts and a model generation, and asked to rate each pair on \textit{language quality}, \textit{scenario plausibility}, \textit{coverage of given concepts}, and an \textit{overall score}, in the Likert scale: \textit{Agree}, \textit{Neutral}, and \textit{Disagree}. Each pair is rated by 3 workers.
% Each pair is rated by 3 workersleading to an average inter-rater agreement (measured by Fleiss' Kappa \cite{fleiss1971mns}) of $0.28$, indicating a \textit{fair} agreement.\footnote{A relatively low agreement score is expected due to the nature of open-ended text generation \cite{clark-etal-2021-thats}.}

%\vspace*{-9mm}
\paragraph{Results.}
Table~\ref{tab:comGen_result_decode} compares different constrained decoding methods on top of the finetuned and off-the-shelf GPT-2, in supervised and zero-shot settings respectively. The key observations are:
\begin{enumerate}[wide, labelwidth=!,listparindent=0pt, labelindent=0pt,noitemsep,topsep=0pt,parsep=2pt,leftmargin =*]
    \item  \methodshort outperforms all previous constrained-decoding methods in both supervised and zero-shot settings. Surprisingly, \textit{unsupervised} \methodshort outperforms all supervised methods based on human evaluation.
    \item  Compared to vanilla \neurologic, \methodshort improves the generation quality while maintaining high constraint satisfaction. The difference is especially substantial in the zero-shot case, where 
    there is more room for incorporating constraint-driven signals 
    % the search problem is much harder 
    due to the lack of supervision
    and the large output space. 
    \item \methodshort reaches similar performance with different lookahead strategies, among which beam lookahead slightly outperforms the others based on human evaluation, and greedy lookahead has the lowest runtime.
\end{enumerate}

\definecolor{brown}{HTML}{9F2B00}
\definecolor{blue}{HTML}{0000D1}
\begin{figure}[t]
\centering
    \includegraphics[width=\linewidth]{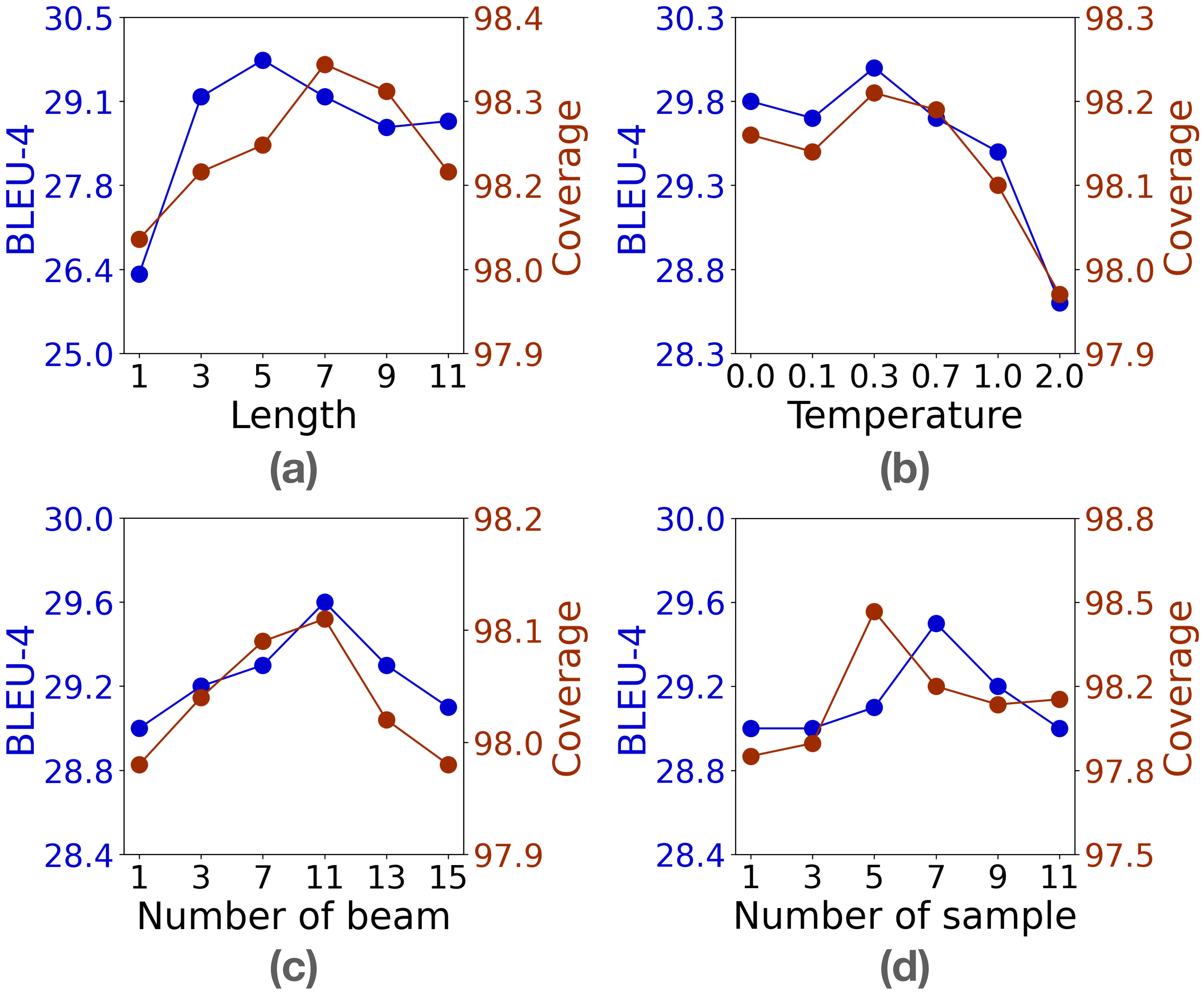}
    \vspace*{-7mm}
     \caption{
     Performance (y-axis) of supervised GPT-2 in terms of \textbf{\textcolor{blue}{BLEU-4}} and \textbf{\textcolor{brown}{Coverage}} with varying look-ahead parameters (x-axis) on \textsc{CommonGen} validation set.}
     \label{fig:comGen_ablation}
\end{figure} 
%\subsubsection{Results III: Ablation}
% \ximing{TODO: add intuition for each hypers}

% \paragraph{Does lookahead improve search?}

% whether increasing the lookahead improves our lookahead heuristics return better approximate solutions than existing decoding algorithms.

\paragraph{Studying lookahead strategies.} 
With an infinite lookahead length $\lk$ and number of lookaheads $|\mathcal{L}_\lk|$, lookahead decoding exactly solves \autoref{eqn:update2}. 
For practical choices of $\lk$ and $|\mathcal{L}_\lk|$,
we empirically study how varying the lookahead strategy and hyperparameters affects performance. %
In \autoref{fig:comGen_ablation}, we study the greedy, soft, beam, and sampling lookahead strategies (\S\ref{ssec:decoding}).

Figure~\ref{fig:comGen_ablation}(a) shows the effect of increasing the lookahead horizon $\ell$ for the greedy strategy.
Increasing the horizon improves up to one point -- e.g., 5-7 steps -- then decreases thereafter, likely due to the difficulty of long-horizon approximation.

Figure~\ref{fig:comGen_ablation}(b) studies the temperature in the soft lookahead, showing that greedy ($\tau=0.0$) performs well, with slight gains if $\tau$ is carefully selected. 
The results suggest that one can safely bypass tuning $\tau$ using fast, greedy lookahead.
%$\tau\rightarrow $

Next, Figure~\ref{fig:comGen_ablation}(c) shows that with beam lookahead, increasing the beam width improves performance up to a certain point (here, 11). 
Similarly, increasing the number of samples with sampling lookahead improves over a single sample, and then reaches an inflection point (Figure~\ref{fig:comGen_ablation}(d)).

% \paragraph{lookahead step}
% Figure~\ref{fig:comGen_ablation}(a) demonstrates the effect of lookahead step in greedy A$^\textbf{*}$ estimation. 
% \paragraph{fusion temperature} Figure~\ref{fig:comGen_ablation}(b) demonstrates the effect of fusion temperature in greedy A$^\textbf{*}$ estimation. 
% \paragraph{lookahead width}
%  demonstrates the effect of lookahead beam size in beam search A$^\textbf{*}$ estimation. ) demonstrates the effect of lookahead sample size in sampling A$^\textbf{*}$ estimation. 

% \rowan{can we compare our predicted lookahead probabiility of satisfying constraints to the real probability of doing so?}

\subsection{\mtfull}
\label{sec:mt}

\begin{table}[t]
\scriptsize
\renewcommand{\arraystretch}{1.1}
    \centering
        \scalebox{1.13}{
    \begin{tabular}{@{} l | c @{\hspace{0.8\tabcolsep}}  c | c  @{\hspace{0.8\tabcolsep}} c @{} }
\specialrule{\heavyrulewidth}{-\heavyrulewidth}{0pt}
\multirow{2}{*} {\hspace{8mm} \textbf{Method}} & \multicolumn{2}{c|}{\citeauthor{dinu-etal-2019-training}} & \multicolumn{2}{c}{Marian MT}\\ 
 &   \textbf{BLUE} &  \textbf{Term\%} & \textbf{BLUE} & \textbf{Term\%}\\ 
\specialrule{\lightrulewidth}{-\lightrulewidth}{0pt}
 Unconstrained  & 25.8 & 76.3 & 32.9 & 85.0\\
 
\rowcolor[gray]{0.90} train-by-app. & 26.0 & 92.9 & -- & --\\
\rowcolor[gray]{0.90}train-by-rep. & 26.0 & 94.5 & -- & --\\

\citet{post-vilar-2018-fast} & 25.3 & 82.0 & 33.0 & 94.3 \\
\neurologic & 26.5 & 95.1 & 33.4 & \underline{97.1} \\
\methodshort (greedy) & \textbf{26.7} & \textbf{95.8}  & \textbf{33.7} & \textbf{97.2} \\
\methodshort (sample) & \underline{26.6} & \underline{95.4} & \textbf{33.7} & \textbf{97.2} \\
\methodshort (beam) & \underline{26.6} & \textbf{95.8} & \underline{33.6} & \textbf{97.2} \\
\specialrule{\heavyrulewidth}{-\heavyrulewidth}{0pt}
    \end{tabular}}
    \caption{Results on constrained machine translation.
    The left section uses the same two-layer transformer model as \citet{dinu-etal-2019-training} for fair comparisons. The right one decodes a stronger Marian MT EN-DE model. The highlighted methods modify training data specifically for constrained decoding, and thus cannot be applied to off-the-shelf models. The best numbers are \textbf{bolded} and the second best ones are \underline{underlined}.
    }  % 
    \label{tab:MT_result}
\end{table}
\begin{table}[t]
\scriptsize
\renewcommand{\arraystretch}{1.1}
\setlength{\tabcolsep}{7.2pt}
    \centering
        \scalebox{1.09}{
    \begin{tabular}{ c | c | l| c  c }
     \specialrule{\heavyrulewidth}{-\heavyrulewidth}{0pt}
 \textbf{\# T} & \textbf{\# Sents.} & \textbf{\hspace{4pt}Decode Method} &   \textbf{BLEU}   & \textbf{Term\%}\\ 
\specialrule{\lightrulewidth}{-\lightrulewidth}{0pt}
\multirow{3}{*}{1} & \multirow{3}{*}{378} & Beam search & 25.4 & 79.6\\
& & \textsc{NeuroLogic} & \underline{26.2} & \underline{95.2} \\
& & \methodshort & \textbf{26.3} & \textbf{95.8} \\
\specialrule{\lightrulewidth}{-\lightrulewidth}{0pt}
\multirow{3}{*}{2+} & \multirow{3}{*}{36} & Beam search & 28.1 & 85.0 \\
& & \textsc{NeuroLogic} & \underline{28.9} & \underline{93.7} \\
& & \methodshort & \textbf{29.3} & \textbf{96.5} \\
\specialrule{\heavyrulewidth}{-\heavyrulewidth}{0pt}
    \end{tabular}}
    \caption{\mtfull performance broken down by the number of constraint terms (\# T). All configurations use the two-layer tranformer from \citet{dinu-etal-2019-training}. The best numbers are \textbf{bolded} and the second best ones are \underline{underlined}.
    }  % 
    \label{tab:MT_ablation}
\end{table}
%\jungo{Ximing, please check the following paragraph.}
It is often critical to have control over machine translation output.
For example, domain-specific dictionaries can be incorporated to force a model to use certain terminology \cite{post-vilar-2018-fast, dinu-etal-2019-training}.
To achieve this goal, much recent work proposed constrained decoding algorithms \interalia{chatterjee-etal-2017-guiding, hokamp-liu-2017-lexically, hasler-etal-2018-neural, hu-etal-2019-improved} or specialized training \cite{dinu-etal-2019-training}.
We demonstrate that \methodshort can be readily applied to off-the-shelf MT systems for constrained machine translation. Specifically, we follow the setup in \citet{dinu-etal-2019-training} and evaluate our method on the WMT17 EN-DE test data \cite{bojar-etal-2017-findings}. The constraint here is to integrate a given custom terminology into the translation output; constraint terms are automatically created from the IATE EU terminology database for 414 test sentences.\footnote{\url{https://github.com/mtresearcher/terminology_dataset}.} 

\paragraph{Approach, Baselines, and Metrics.} We experiment with two MT systems: \citeauthor{dinu-etal-2019-training} (two-layer transformer) and the off-the-shelf Marian MT \cite{mariannmt}. We compare with previous constrained decoding algorithms, including DBA \cite{post-vilar-2018-fast}, \neurologicshort \cite{lu-etal-2021-neurologic} and also specialized training proposed by \citet{dinu-etal-2019-training}. Following \citet{dinu-etal-2019-training}, we report BLEU scores and term use rates, computed as the percentage of times a given constraint term was generated in the output out of the total number of constraint terms.

\paragraph{Results.} Table \ref{tab:MT_result} presents experimental results with \citeauthor{dinu-etal-2019-training}'s model and Marian MT.
We can see that in either case, \methodshort outperforms all prior methods both in BLEU and term coverage. Besides better generation quality and constraint coverage, \methodshort also benefits from its plug-and-play flexibility with any off-the-shelf MT system compared to previous training-based methods. Table \ref{tab:MT_ablation} breaks down the model performance by the number of constraint terms.
We see that \methodshort improves upon the others, especially when the constraint is complex with multiple constraint terms. (\eg 96.5 vs.\ 93.7 from \textsc{NeuroLogic} in term coverage).

%\ximing{TODO: some writing below is unfinished yet}

\subsection{\tttfull}
\label{sec:table-to-text}
%\jungo{Edited this section. Please check.} \liwei{Look great to me!}
The table-to-text task aims to generate natural language text conditioned on structured table data; their applications include automatic generation of weather/sports reports \cite{liang-etal-2009-learning, wiseman-etal-2017-challenges} or dialogue responses \cite{wen-etal-2016-multi}.
Constrained generation algorithms can be used to ensure that the output text is
% factual and 
consistent with the input structured data.
% Despite the remarkable progress neural language models have made on this task, their performance strongly relies on large amounts of domain-specific annotated data, limiting their applicability in real-world applications.
We follow the few-shot setup of 
% \citet{su2021few}, we explore few-shot learning settings for table-to-text generation.
% Specifically, we follow the set up in 
~\citet{chen-etal-2020-kgpt} on the E2ENLG \cite{dusek.etal2018:inlg} dataset, where we use randomly-sampled 0.1\%, 0.5\%, 1\%, 5\% of training instances for finetuning. 
% The task is to generate a restaurant review that incorporates all the information in the given table.

\paragraph{Approach, Baselines, and Metrics.} 
Following \citet{shen-etal-2019-pragmatically}, we linearize the given table into a string and finetune GPT-2 with given few-shot examples.
We first compare \methodshort with three previous constrained decoding algorithms: CBS \cite{anderson-etal-2017-guided}, GBS \cite{hokamp-liu-2017-lexically}, and \neurologicshort \cite{lu-etal-2021-neurologic}, based on few-shot GPT-2 finetuned with 0.1\% data.
Then we compare our approach, \methodshort on top of GPT-2, with previous table-to-text methods, including TGen \cite{dusek-jurcicek-2016-sequence}, Template-GPT-2 \cite{chen-etal-2020-logical}, KGPT \cite{chen-etal-2020-kgpt}, in multiple few-shot settings with various numbers of training instances.
We report standard automatic metrics used in the E2ENLG challenge, 
% NIST \cite{lin-hovy-2003-automatic}, 
% BLEU \cite{papineni2002bleu}, METEOR \citep{banerjee2005meteor}, CIDEr \citep{vedantam2015cider}, ROUGE  \cite{lin2004rouge}, 
as well as information coverage-- the average percentage of given information that is present in the generation. 
\begin{table}[t]
\footnotesize
\renewcommand{\arraystretch}{1.1}
\setlength{\tabcolsep}{1.6pt}
    \centering
    \resizebox{\linewidth}{!}{%
    \begin{tabular}{l |c c c c c c }
     \specialrule{\heavyrulewidth}{-\heavyrulewidth}{0pt}
 \hspace{18pt}\textbf{Decode Method} & \textbf{NIST} & \textbf{BLEU} & \textbf{METEOR}   & \textbf{CIDEr}  & \textbf{ROUGE} & \textbf{Coverage}\\ 
 \specialrule{\lightrulewidth}{-\lightrulewidth}{0pt}
 Beam Search & 3.82 & 42.8 & 32.6 & 10.8 & 57.8 & 73.6\phantom{.}\\
\textsc{CBS} & 6.50 & 42.3 & 36.4 & 13.0 & 54.3 & 91.6\phantom{.}\\
\textsc{GBS} & 6.26 & 40.7 & 36.7 & 12.9 & 54.2 & 94.1\phantom{.} \\
\textsc{NeuroLogic} & 6.95 & 47.6 & 38.9 & 16.3 & 58.7 & 97.6\phantom{.} \\
\specialrule{\lightrulewidth}{-\lightrulewidth}{0pt}
\methodshort (greedy) & \textbf{7.11} & \underline{49.2} & \underline{40.0} & \textbf{17.5} & \textbf{60.0} & \textbf{100.0} \\
\methodshort (beam) & \underline{7.01} & 48.9 & \underline{40.0} & \underline{17.2} & \underline{59.8} & \underline{99.9}\phantom{.} \\
\methodshort (sample) & \textbf{7.11} & \textbf{49.3} & \textbf{40.1} & \textbf{17.5}  & \textbf{60.0} & \textbf{100.0} \\
\specialrule{\heavyrulewidth}{-\heavyrulewidth}{0pt}
    \end{tabular}}
    \caption{Performance of different decoding methods with few-shot GPT-2 finetuned on 0.1\% E2ENLG data. The best numbers are \textbf{bolded} and the second best ones are \underline{underlined}. }  % 
    \label{tab:t2t_result}
\end{table}
\begin{table}[t]
\scriptsize
\renewcommand{\arraystretch}{1.1}
\setlength{\tabcolsep}{3.7pt}
    \centering
    \resizebox{\linewidth}{!}{
    \begin{tabular}{l |c c c c }
     \specialrule{\heavyrulewidth}{-\heavyrulewidth}{0pt}
         \hspace{41pt}\textbf{Method} & \textbf{0.1\%} & \textbf{0.5\%} &  \textbf{1\%}   & \textbf{5\%}\\ 
         \specialrule{\lightrulewidth}{-\lightrulewidth}{0pt}
        TGen \cite{dusek-jurcicek-2016-sequence} & 3.6 \phantom{0} & 27.9 & 35.2 & 57.3  \\
        Template-GPT-2 \cite{chen-etal-2020-logical} & 22.5 & 47.8 & 53.3 & 59.9 \\
        KGPT-Graph \cite{chen-etal-2020-kgpt} & 39.8 & 53.3 & 55.1 & 61.5  \\
        KGPT-Seq  \cite{chen-etal-2020-kgpt} & 40.2 & 53.0 & 54.1 & 61.1  \\
        \specialrule{\lightrulewidth}{-\lightrulewidth}{0pt}
        GPT-2 & 42.8 & \underline{57.1} & 56.8 & 61.1 \\
        GPT-2 + \neurologicshort & \underline{47.6} & 56.9 & \underline{58.0} & \underline{62.9}  \\
        GPT-2 + \methodshort  (greedy) & \textbf{49.2} & \textbf{58.0} & \textbf{58.4} & \textbf{63.4}\\
    \specialrule{\heavyrulewidth}{-\heavyrulewidth}{0pt}
    \end{tabular}
    }
    \caption{Few-shot results (BLEU-4) on E2ENLG test set with 0.1\%, 0.5\%, 1\%, 5\% of training instances. The best numbers are \textbf{bolded} and the second best ones are \underline{underlined}.}  % 
    \label{tab:t2t_ablation}
\end{table}

\definecolor{blue1}{HTML}{1F77B4}
\definecolor{purple1}{HTML}{9E0D9E}
\begin{figure}[t]
\centering
    \includegraphics[width=1\linewidth]{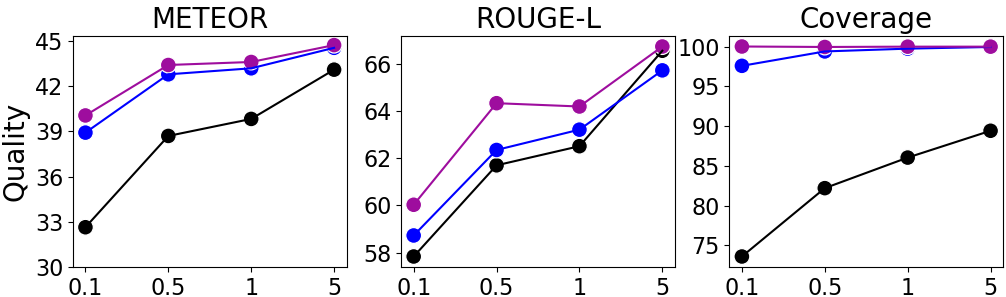}
    \vspace*{-7mm}
     \caption{
     Performance (y-axis) of supervised GPT-2 on E2ENLG, with a varying amount of training data for supervision (x-axis). The \textbf{\textcolor{purple1}{purple}}, \textbf{\textcolor{blue}{blue}}, and \textbf{\textcolor{black}{black}} line denote decoding with \methodshort, \textsc{NeuroLogic} and conventional beam search respectively. 
     }
     \label{fig:ablation_t2t_amount}
\end{figure} 

%, which leverages either specialized model architecture or additional pretraining on massive table-to-text corpus
\paragraph{Results.} Table \ref{tab:t2t_result} presents results from varying decoding algorithms based on few-shot GPT-2 finetuned with 0.1\% of the data. 
\methodshort substantially outperforms all previous methods with respect to all metrics; it consistently improves generation quality while achieving (almost) perfect constraint satisfaction. Previous work, like CBS and GBS, improves constraint satisfaction, but negatively affects the text quality, as indicated by drops in BLEU and ROUGE. Table~\ref{tab:t2t_ablation} compares \methodshort on top of GPT-2 with previous table-to-text approaches. As before, \methodshort outperforms all prior approaches by a large margin, even if the latter ones leverage either specialized model architecture or additional pretraining on massive table-to-text corpora. Additionally, Figure~\ref{fig:ablation_t2t_amount} compares the performance (y-axis) of few-shot GPT-2 with \methodshort (\textbf{\textcolor{purple1}{ purple line}}), \neurologicshort (\textbf{\textcolor{blue}{blue line}}), and conventional beam search (\textbf{\textcolor{black}{black line}}) as a function of the varying amount of training instances (x-axis). We find the relative gain brought by \methodshort increases as we reduce the amount of few-shot examples. Results above demonstrate the promise of decoding algorithms to address unsatisfying performance in few-shot scenarios due to insufficient learning.

%Seen in Table \ref{tab:t2t_result} are results from varying decoding algorithms.
%We see that the \methodshort methods substantially outperform the previous methods in all automatic metrics and term coverage; in particular, \methodshort greedy and sample are both able to satisfy all constraints.
%demonstrate the promise of decoding algorithms to address the unsatisfying performance and factual inconsistency due to the insufficient learning.
%\liwei{I assume this part is unfinished? Can we elaborate the results, at least mention which tables are we looking at?}
%\jungo{Edited. Please check.} \liwei{Looks great to me! Should we talk about Table 5 somewhere as well?} \ximing{TODO}

\newcolumntype{x}[1]{%
>{\centering\hspace{0pt}}p{#1}}%
\newcolumntype{?}{!{\vrule width 1pt}}
\renewcommand{\arraystretch}{1.1}
\begin{table*}[t]
\setlength{\tabcolsep}{3.7pt}
    \centering
    \resizebox{\linewidth}{!}{
    \begin{tabular}{p{5.0cm}| c c c c c  c |c c c c}
     \specialrule{1.5pt}{-1.5pt}{0pt}
 \multirow{2}{*} {\hspace{13mm}\textbf{Decode Method}}  & \multicolumn{6}{c}{\textbf{Automatic Evaluation}}  &  \multicolumn{4}{|c}{\textbf{Human Evaluation}} \\ 
 & \textbf{ROUGE} &  \textbf{BLEU}  & \textbf{METEOR} &\textbf{CIDEr} & \textbf{SPICE} &\textbf{Coverage} & \textbf{Grammar} & \textbf{Fluency} & \textbf{Meaningfulness} & \textbf{Overall} \\ 
\specialrule{1.0pt}{-1.0pt}{0pt}
CGMH \cite{miao2019cgmh} & 28.8 & 2.0\phantom{.} & 18.0 & 5.5\phantom{.} & 21.5 & 18.3\phantom{.} & 2.28 & 2.34 & 2.11 & 2.02 \\
TSMH \cite{zhang-etal-2020-language-generation} & 42.0 & 4.3\phantom{.} & 25.9 & 10.4 & 37.7 & \underline{92.7}\phantom{.} & 2.35 & 2.28 & 2.37 & 2.22 \\
\textsc{NeuroLogic} \cite{lu-etal-2021-neurologic}  & 38.8 & 11.2 & 24.5 & 18.0 & 41.7 & 90.6\phantom{.} & 2.78 & 2.71 & 2.49 & 2.51 \\
\specialrule{1.0pt}{-1.0pt}{0pt}
\methodshort (greedy) & \textbf{43.7} & \textbf{14.7} & \underline{28.0} & \textbf{20.9} & \underline{47.7} & \textbf{100.0} & \textbf{2.83} & \underline{2.77} & \underline{2.74} & \textbf{2.76} \\
\methodshort (beam) & 42.9 & 14.4 & 27.8 & 20.3 & 46.9 & \textbf{100.0} & \underline{2.81} & \textbf{2.86} & \textbf{2.76} & \underline{2.75} \\
\methodshort (sample) & \underline{43.5} & \underline{14.6} & \textbf{28.2} & \underline{20.8} & \textbf{47.8} & \textbf{100.0} & \textbf{2.83} & 2.75 & \textbf{2.76} & 2.73 \\
    \specialrule{1.5pt}{-1.5pt}{0pt}
    \end{tabular}}
    \caption{Performance of different unsupervised decoding algorithms on interrogative question generation.}  % 
    \label{tab:result_question}

\end{table*}
\subsection{\questiongenfull}
\label{sec:interrogative}
Despite the success of supervised techniques in natural language generation, it needs to be trained with massive task-specific data, which is non-trivial to acquire. We investigate a zero-shot text generation task proposed by \citet{zhang-etal-2020-language-generation}: constrained question generation, where no training data is available. Given a set of keywords (e.g., {Nevada, desert, border}), the task is to use an off-the-self language model to generate an interrogative question containing given keywords (e.g., ``What is the name of the desert near the border of Nevada?'').
Two types of constraints are enforced for this task: 1) keyword constraints - the output question must include all the keywords provided, and 2) syntactic constraints - the output question must be in the interrogative form, the first word must be \emph{wh-} question words, and the second or third word must be auxiliary verbs or copula words. 

\paragraph{Approach, Baselines, and Metrics.} 
We leverage off-the-shelf language model GPT-2 and compare \methodshort with three previous constrained decoding methods, CGMH \cite{miao2019cgmh}, TSMH \cite{zhang-etal-2020-language-generation} and \neurologicshort \cite{lu-etal-2021-neurologic}. CGMH and TSMH are two Metropolis-Hastings sampling-based decoding algorithms that have shown strong performance in unsupervised constrained generation. For automatic evaluation, we report standard generation metrics and keyword Coverage similar to previous task \commongen.
For the human evaluation, we sample 100 test examples and employ workers from AMT to evaluate the generated interrogative questions. Workers are given a set of keywords and model generation. They are asked to evaluate the generation based on 3 individual qualities (\ie grammar, fluency, meaningfulness) and provide an overall quality score, using the 3-point Likert scale. Each example is averaged across 3 workers. We include the human evaluation template in Figure \ref{fig:question_gen_human_eval} of the Appendix \ref{sec:appendix:human_eval}. 

\paragraph{Results.} Table~\ref{tab:result_question} presents comparisons across different decoding methods based on off-the-shelf language models. We can see that \methodshort outperforms all previous methods with respect to both automatic and manual metrics; it remarkably enhances the generation quality while achieves perfect constraint satisfaction. The difference between \neurologicshort and \methodshort is particularly large compared to other tasks. The search problem is much harder here, due to the lack of supervision and complex logical constraint involving both keywords and syntax. Results above demonstrate the effectiveness of \methodshort in tackling more challenging constrained generation problems.

%\subsection{Template Infilling}

\newcolumntype{x}[1]{%
>{\centering\hspace{0pt}}p{#1}}%
\newcolumntype{?}{!{\vrule width 1pt}}
\renewcommand{\arraystretch}{1.1}
\begin{table*}[t]
\setlength{\tabcolsep}{3.2pt}
    \centering
    \resizebox{\linewidth}{!}{
    \begin{tabular}{l | c c c | c c |c c c c c}
     \specialrule{1.5pt}{-1.5pt}{0pt}
    \multirow{2}{*}{\hspace{58pt}\textbf{Decode Method}} & \multicolumn{3}{c|}{\textbf{Fluency}} & \multicolumn{2}{c|}{\textbf{Diversity}} & \multicolumn{5}{c}{\textbf{Human Eval}}\\
  & \textbf{PPL} &  \textbf{BLEU-1}  & \textbf{BLEU-2} & \textbf{Uniq. 3-gram} &\textbf{Uniq. 4-gram} & \textbf{Grammar} & \textbf{Fluency} & \textbf{Coherence} &  \textbf{Interest} & \textbf{Overall} \\ 
  \specialrule{1.0pt}{-1.0pt}{0pt}
beam search & 2.24 & 33.7 & 16.5 & 34.09k & 41.91k & 2.81 & 2.50 & 2.46 & 2.27 & 2.32 \\
beam search + \astaresque (greedy) & \textbf{2.11} & \underline{34.3} & \underline{16.7} & 34.94k & 43.02k & \textbf{2.94} & \underline{2.71} & 2.56 & 2.50 & \underline{2.57} \\
beam search +  \astaresque (beam) & \underline{2.14} & \textbf{34.4} & \textbf{16.8} & \underline{35.03k} & \underline{43.12k} & \textbf{2.94} & \textbf{2.72} & \textbf{2.62} & \textbf{2.61} & \textbf{2.63} \\
beam search +  \astaresque (sample)& 2.16 & \textbf{34.4} & \underline{16.7} & \textbf{35.41k} & \textbf{43.64k} & \underline{2.92} & \underline{2.71} & \underline{2.59} & \underline{2.52} & \underline{2.57} \\
   \specialrule{1.0pt}{-1.0pt}{0pt}
top-k sample & 4.01 & 31.4 & 13.9 & \underline{48.36k} & \underline{56.62k} & 2.69 & 2.38 & 2.23 & 2.30 & 2.15 \\
top-k sample +  \astaresque (greedy) & \textbf{3.68} & \underline{32.1} & \underline{14.3} & \textbf{48.44k} & \textbf{56.63k} & \textbf{2.88} & \textbf{2.57} & \textbf{2.48} & \textbf{2.49} & \textbf{2.47} \\
top-k sample +  \astaresque  (beam) & 3.75 & \textbf{32.2} & \textbf{14.4} & 48.27k & 56.36k & \underline{2.84} & 2.49 & 2.39 & 2.40 & 2.34 \\
top-k sample +  \astaresque (sample)& \underline{3.70} & 32.0 & 14.2 & 48.04k & 56.15k & \underline{2.84} & \underline{2.55} & \underline{2.47} & \underline{2.48} & \underline{2.44} \\
    \specialrule{1.5pt}{-1.5pt}{0pt}
    \end{tabular}}
    \caption{Performance of different decoding algorithms on RocStories test set.}  % 
    \label{tab:results_rocstories}

\end{table*}

\definecolor{blue1}{HTML}{1F77B4}
\definecolor{purple1}{HTML}{9E0D9E}
\definecolor{brown}{HTML}{9F2B00}
\definecolor{green1}{HTML}{0F8F00}
\begin{figure}[t]
\centering
    \includegraphics[width=\linewidth]{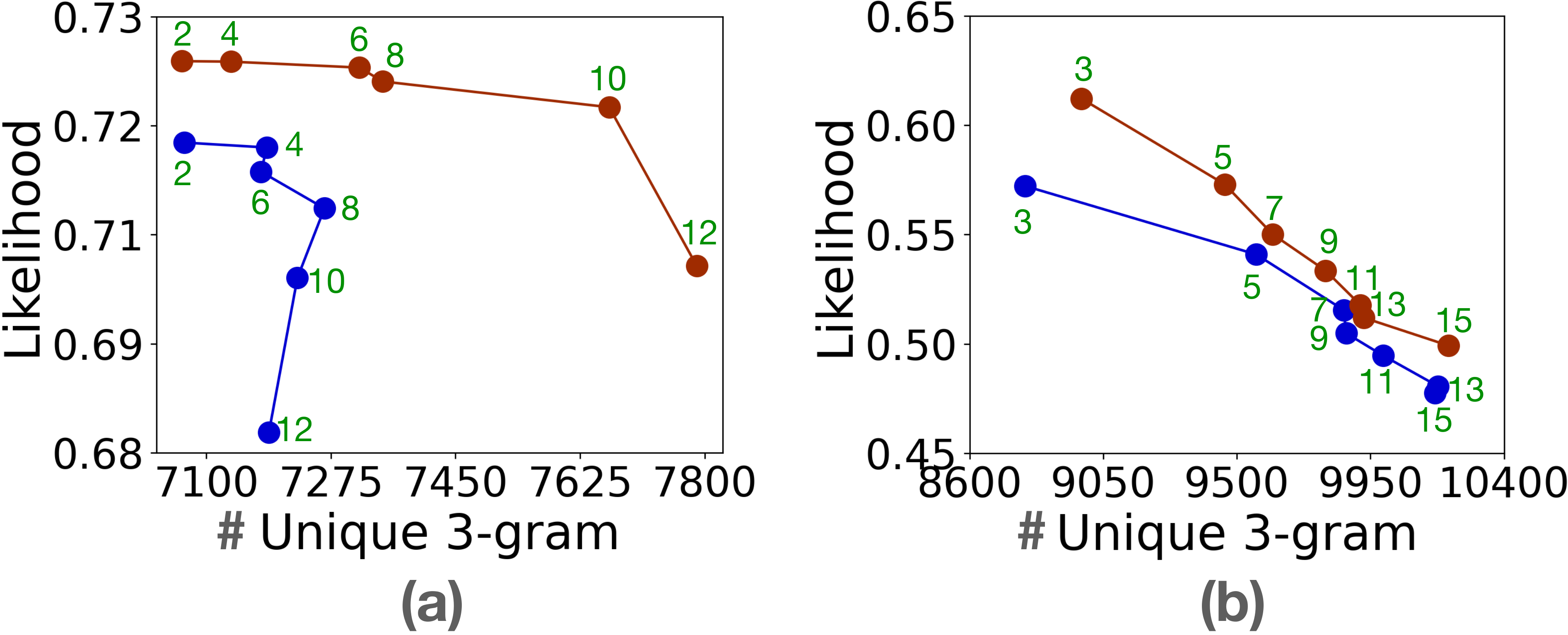}
    \vspace*{-7mm}
     \caption{
     Likelihood (y-axis) vs. number of unique 3-grams (x-axis) using supervised GPT-2 on RocStories. Figure \textbf{(a)} denotes decoding with beam search, with a varying amount of \textcolor{green1}{beam size}. Figure \textbf{(b)} denotes decoding with top-k sampling, with a varying amount of \textcolor{green1}{k value}. The \textbf{\textcolor{brown}{brown}} and \textbf{\textcolor{blue}{blue}} line denotes with and without \astaresque heuristics separately. 
     }
     \label{fig:roc_tradeoff}
\end{figure} 

\section{Experiments: Unconstrained Generation}
So far we have experimented with constrained text generation, but here we demonstrate that 
% \methodshort 
\methodshort decoding can also improve \textit{unconstrained} generation.
Specifically, we investigate whether \astaresque decoding with our unconstrained lookahead heuristic (\autoref{eqn:heur-unc}) can (i) 
% improve beam search in an open-ended generation setting,
improve beam search, which typically struggles in open-ended settings
% eliminate degeneracies that occur with beam search in open-ended settings 
\cite{holtzman2020curious,welleck2020neural}, 
(ii) improve \textit{sampling} algorithms that are commonly used in open-ended generation.

\subsection{\storygenfull}
% \peter{FROM NOAH: this section seems redundant with earlier parts of the paper (see his note)}
% Open-ended text generation such as storytelling is an intuitive yet challenging task. Despite the encouraging progress of neural language models, it remains an open
% question what the best decoding strategy is to decode from a language model. Maximization-based decoding methods such as beam search is known to be flawed - output text is repetitive or dull, with high frequency tokens used too often and interesting content words used too rarely \cite{holtzman2020curious, welleck2020neural}. Sampling-based decoding methods such as top-k and nucleus sampling decrease the repetitiveness of texts at the expense of verifiability. \cite{massarelli-etal-2020-decoding}. In both cases, decoding proceeds auto-regressively conditioned on history sequence to maximize or sample from likelihood. We investigate whether incorporating future information through \astar looking ahead would enable beam search or sampling draw better from language model. Specially, 
We investigate story generation with RocStories \cite{mostafazadeh-etal-2016-corpus}. Given the first sentence as a prompt $\xb$, the task is to generate the rest of story continuation $\yb$. 

\paragraph{Approach, Baselines and Metrics.} We consider storytelling as a
conditional generation task, and finetune GPT-2 as a sequence-to-sequence model. 

We apply \astaresque decoding with our unconstrained lookahead heuristic (\autoref{eqn:heur-unc}) to (i) beam search, the setting used so far in the experiments, and (ii) top-k sampling \cite{fan-etal-2018-hierarchical}, a commonly used sampling algorithm in open-ended generation. 
For top-k sampling, we use the heuristic to adjust the probability scores, then renormalize.

For automatic evaluation, besides commonly used automatic metrics for storytelling, including perplexity and BLEU, we also report unique n-grams as a measure for diversity. 
%We report commonly used automatic metrics for storytelling: 1) Perplexity: smaller perplexity scores indicate better fluency in general. 2) BLEU. However for open-ended generation, BLEU scores will become extremely low for large n, since there can be multiple plausible output given the same input but only one is given as the groundtruth. Following \citet{Guan2020AKP}, we report BLEU with $n = 1,2$ 3) Unique n-gram: 
For the human evaluation, we sample 100 stories from the test set and we employ workers from AMT to evaluate the model generations. Workers are given the first sentence of the story (i.e., prompt), and the model-generated continuation of the story. They are asked to evaluate the continuation of the story on 4 individual qualities (i.e., grammar, fluency, story flow, interestingness) and provide an overall quality score, using the 3-point Likert scale. Each example is averaged across 3 workers. We include the human evaluation template in Figure \ref{fig:rocstories_human_eval} of the Appendix \ref{sec:appendix:human_eval}.

% Additionally, we conduct human evaluation on 100 test examples with workers from Amazon Mechanical Turk. We include our evaluation template in the appendix. Workers are presented a pair of concepts and model generation, and asked to rate each pair on language quality, scenario plausibility, coverage of given concepts, and overall score, using the 3-point Likert scale \footnote{The choices presented to wokers are \textit{Agree}, \textit{Neutral}, \textit{Disagree}}. Each pair is rated by 3 workers
\paragraph{Results.} Table~\ref{tab:results_rocstories} presents the results of beam search and top-k sampling with and without \astaresque heuristics. We can see that \astaresque heuristics enable both beam search and top-k sampling to generate more fluent, coherent and interesting stories. 
For beam search, our \astaresque heuristic not only enhances generation quality-- e.g. improving human evaluation scores from 2.32 to 2.63--but also boosts generation diversity, as reflected by the number of unique n-grams. 
For top-k sampling, \astar heuristics also improves generation quality, while maintaining comparable diversity. We notice that beam lookahead works the best for beam search, and greedy lookahead works the best for top-k sampling. 
We suspect that beam lookahead gives the most accurate estimate of the future path that beam search is likely to reach, while the greedy lookahead provides an estimate that is lower than what obtained by beam search, which may better resemble a continuation from top-$k$ sampling. %the future path that has the highest probability top-k-sampling would sample from. 

\paragraph{Ablations.}  We study the effect of \astaresque decoding with different decoding hyperparameters: beam size in beam search and k value in top-k sampling. Figure~\ref{fig:roc_tradeoff} plots the fluency (measured by likelihood) versus diversity (measured by unique 3-grams) for generations with various beam sizes or k values. Ideally, we want generations to be both fluent and diverse, centering around the top-right center. However, we observe a fluency and diversity tradeoff in practice. Interestingly, we observe that \astaresque decoding flattens this trend and results in larger area under the curve. The effect is especially obvious for beam search. The results above demonstrate that \astaresque decoding can guide generation towards a more favorable output space that cannot be reached with conventional decoding methods, regardless of decoding hyperparameters. 
\section{Related Work}
\paragraph{\astar search in NLP.}
Many classical NLP problems (\eg parsing, text alignment) can be seen as structured prediction subject to a set of task-specific constraints.
For many such problems, \astar search has been used effectively~\cite{och2001efficient,haghighi-etal-2007-approximate,hopkins2009cube,meister2020best}. 
For example, \citet{klein-manning-2003-parsing,zhang2006efficient,auli-lopez-2011-efficient,lee-etal-2016-global} have used it in the context of parsing. 
Similar approaches are used for finding high-probability alignments~\cite{naim2013text}. 
Despite these applications, applying informed heuristic search to text generation with autoregressive language models has been under-explored, which is the focus of this work.

\paragraph{Decoding strategies for text generation.}
The rise of autoregressive language models like GPT~\cite{radford2018improving} has inspired a flurry of work on decoding strategies~\cite{post-vilar-2018-fast,ippolito2019comparison,zheng2020opportunistic,leblond2021machine, West2021ReflectiveDB}. 
These works often focus on incorporating factors like diversity~\cite{ippolito2019comparison}, fluency~\cite{holtzman2020curious} or constraints~\cite{anderson-etal-2017-guided,hokamp-liu-2017-lexically,post2018fast,miao2019cgmh,welleck2019non,zhang-etal-2020-language-generation,qin2020backpropagation,lu-etal-2021-neurologic}. 
% \youngjae{
%  extend beam search to set hypotheses that satisfy the lexical-constrains to be considered during generation
% Miao et al. (2019)~\cite{miao2019cgmh} proposes a sampling-based conditional generation by Metropolis-Hastings (CGMH), where the constrained words are inserted/deleted/edited by the Metropolis-Hastings scheme.
% Welleck et al. ~\cite{welleck2019non} suggests a tree-based constrained text generation, which recursively generates text from constrained tokens.
% And Zhang et al. ~\cite{zhang-etal-2020-language-generation} proposes tree search enhanced MCMC that handles the combinatorial constraints (TSMH).
%Post and Vilar (2018)~\cite{post2018fast} propose efficient dynamic beam allocation and He et al~\cite{hu-etal-2019-improved} make it GPU efficient. 
Among constrained decoding methods, previous works such as constrained beam search~\cite{anderson-etal-2017-guided} and grid beam search~\cite{hokamp-liu-2017-lexically}, have worked on extending beam search to satisfy lexical constraints during generation.

Other works have focused on the mismatch between monotonic decoding and satisfying constraints that may depend on a full generation. \citet{miao2019cgmh} propose a 
% sampling-based conditional generation by Metropolis-Hastings Sampling (CGMH), 
sampling-based conditional generation method using Metropolis-Hastings sampling (CGMH), where the constrained words are inserted/deleted/edited by the Metropolis-Hastings scheme, allowing a full generation to be edited towards desired properties.
\citet{welleck2019non} develop a tree-based constrained text generation, which recursively generates text in a non-monotonic order given constraint tokens, ensuring constraints are satisfied.
\citet{zhang-etal-2020-language-generation} proposes tree search enhanced MCMC that handles combinatorial constraints (TSMH).
% }
\citet{qin2020backpropagation} instead casts constrained decoding as a continuous optimization problem that permits gradient-based updates. \citet{West2021ReflectiveDB} encodes constraints as generated contexts which models condition on to encourage satisfaction. Compared to these past works, \method explicitly samples future text to estimate viability of different paths towards satisfying constraints. Our approach is based on \citet{lu-etal-2021-neurologic}, which incorporates constraints in Conjunctive Normal Form (CNF), but we extend this into the future with our lookahead heuristics.

\section{Conclusion}
%\jungo{A large part of this looks like an intro. Suggesting we compress this.}
%Neural language generation has progressed rapidly in recent years, largely driven by language models that mostly rely on information from the \textit{past} during text generation.
%These models tend to make locally preferable predictions conditioned only on the \textit{past}, which result in generations that might be globally degenerated.
Inspired by the \astar search algorithm, we introduce \method decoding, 
% which enables traditional \textit{left-to-right} decoding algorithms to use the \textit{future} while making decisions based on \textit{past} context.
which brings A*-like heuristic estimates of the \textit{future} to common \textit{left-to-right} decoding algorithms for neural text generation.
% , while preserving the efficiency demanded by large-scale neural language models.
\method's lookahead heuristics improve over existing decoding methods (\eg \neurologicshort, beam, greedy, sample decoding methods) in both \textit{constrained} and \textit{unconstrained} settings across a wide spectrum of tasks.
%including constrained commonsense generation, machine translation, table-to-text generation, interrogative sentence generation, and open-text story generation.
% Particularly, we introduce \methodnospace, which 
% Our work demonstrates the promises of adding  on top of the existing decoding 
Our work demonstrates the promise of moving beyond the current paradigm of unidirectional decoding for text generation, by taking bidirectional information from both the \textit{past} and \textit{future} into account to generate more globally compatible text.

\section*{Acknowledgment}
This work was supported in part by Natural Sciences and Engineering Research Council of Canada (NSERC) (funding reference
number 401233309), DARPA MCS program through NIWC Pacific (N66001-19-2-4031),  Google Cloud Compute, and Allen Institute for AI, Microsoft PhD Fellowship.
% and the Allen Institute for AI.  % dropping as there are full-time AI2 folks on the author list

\bibliography{anthology,custom}
\bibliographystyle{acl_natbib}

\clearpage

\appendix

% {\Large \bf Supplementary Material }

%\section{Example Appendix}
\label{sec:appendix}

% This is an appendix.

\section{Human Evaluation}
\label{sec:appendix:human_eval}

\begin{figure*}
    \centering
    \includegraphics[width=1\linewidth]{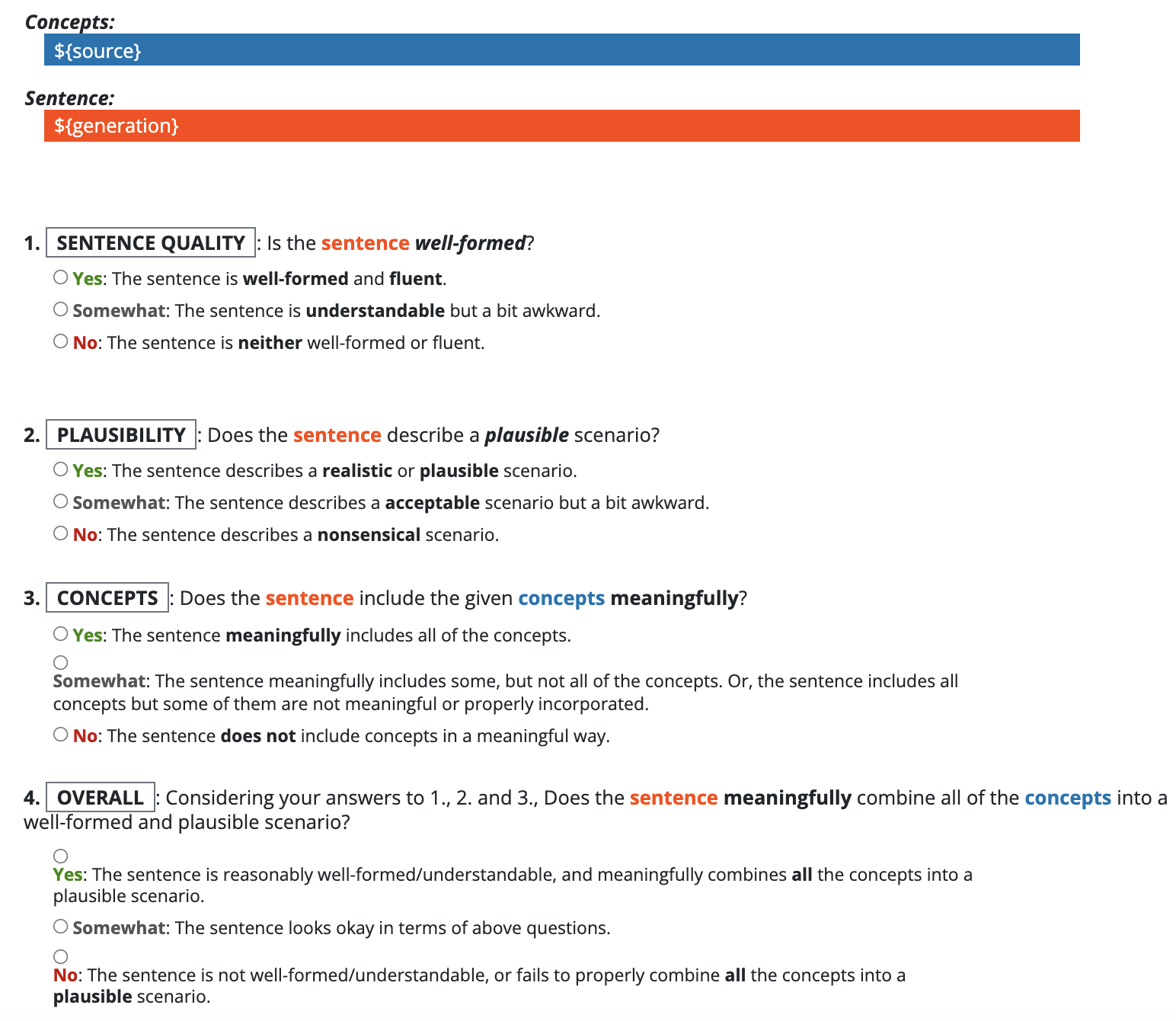}
    \caption{Human evaluation template for the Constrained Commonsense Generation task.}
    \label{fig:commongen_human_eval}
\end{figure*}

\begin{figure*}
    \centering
    \includegraphics[width=1\linewidth]{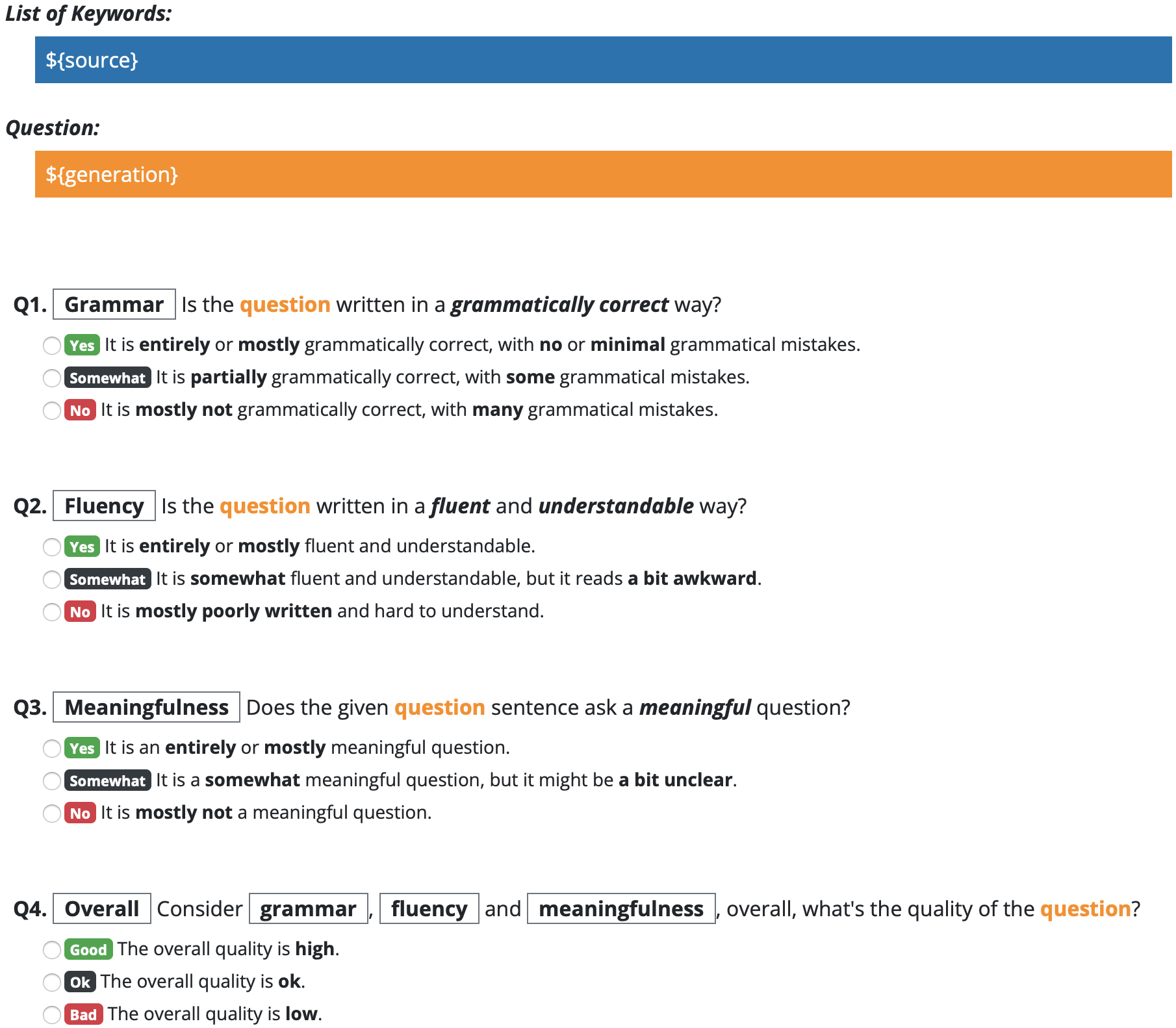}
    \caption{Human evaluation template for the Interrogative Sentence Generation task.}
    \label{fig:question_gen_human_eval}
\end{figure*}

\begin{figure*}
    \centering
    \includegraphics[width=1\linewidth]{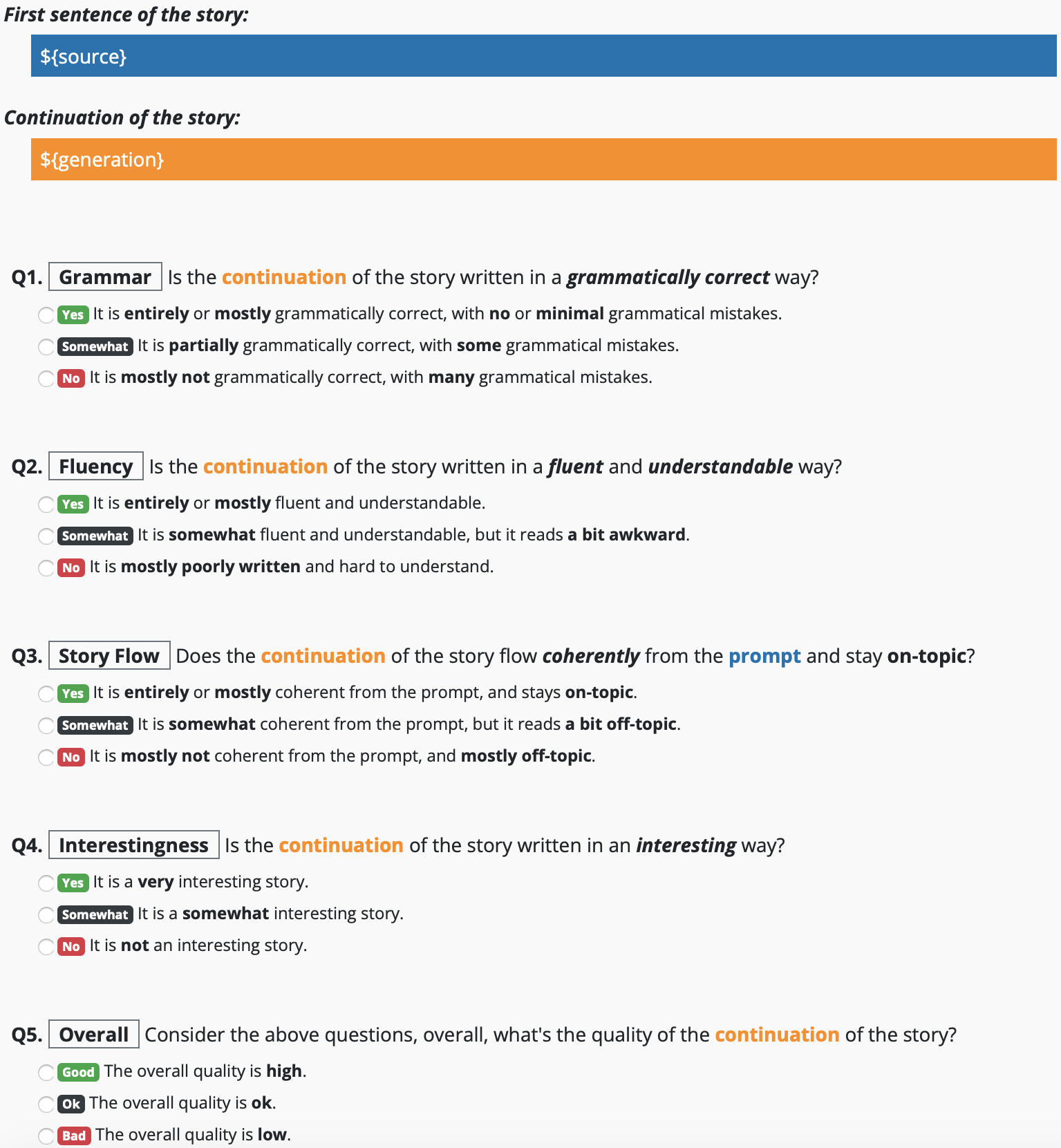}
    \caption{Human evaluation template for the RocStories task.}
    \label{fig:rocstories_human_eval}
\end{figure*}

We include screenshots of the human evaluation templates for CommonGen (Figure \ref{fig:commongen_human_eval}), Interrogative Sentence Generation (Figure \ref{fig:question_gen_human_eval}), and RocStories (Figure \ref{fig:rocstories_human_eval}) tasks.

\end{document}